\pdfoutput=1

\PassOptionsToPackage{sort}{natbib}
\documentclass[11pt]{article}

\usepackage[final]{acl}

\usepackage{times}
\usepackage{latexsym}
\usepackage{multirow}
\usepackage[T1]{fontenc}

\usepackage{tipa}
\usepackage[utf8]{inputenc}
\usepackage{paralist}
\usepackage{enumitem}
\usepackage{microtype}

\usepackage{inconsolata}
\usepackage[most]{tcolorbox} 
\usepackage{mdframed}
\usepackage{xcolor} 
\usepackage{booktabs}
\usepackage{graphicx}
\usepackage[skip=5pt]{caption}

\usepackage{paralist}
\newcommand{\header}[1]{\par \vspace*{1mm}\noindent\textbf{#1}.}

\newcommand{\amin}[1]{\textcolor{black}{#1}}

\definecolor{myblue}{HTML}{0B8EE5}
\definecolor{leidenblue}{HTML}{001158}
\definecolor{myorange}{HTML}{FBD150}

\definecolor{ballblue}{rgb}{0.13, 0.67, 0.8}
\definecolor{caribbeangreen}{rgb}{0.0, 0.8, 0.6}
\definecolor{cherryblossompink}{rgb}{1.0, 0.72, 0.77}
\definecolor{fluorescentorange}{rgb}{1.0, 0.75, 0.0}
\definecolor{skyblue}{rgb}{0.53, 0.81, 0.92}
\definecolor{stildegrainyellow}{rgb}{0.98, 0.85, 0.37}

\newcommand{\highlightgreen}[1]{\colorbox{caribbeangreen!40}{#1}}
\newcommand{\highlightred}[1]{\colorbox{pink}{#1}}

\author{Amin Abolghasemi\textsuperscript{1} \qquad Leif Azzopardi\textsuperscript{2} \qquad
 Seyyed     Hadi Hashemi\textsuperscript{3}\\  \textbf{Maarten de Rijke\textsuperscript{4}} \qquad
\textbf{Suzan Verberne\textsuperscript{1}} \\
  \makebox[.4\linewidth]{\textsuperscript{1}Leiden University, Netherlands}  \makebox[.4\linewidth]{\textsuperscript{2}University of Strathclyde, UK} \\
  \makebox[.4\linewidth]{\textsuperscript{3}eBay Inc., Netherlands}  \makebox[.4\linewidth]{\textsuperscript{4}University of Amsterdam, Netherlands}
   \\
  \texttt{\{m.a.abolghasemi, s.verberne\}@liacs.leidenuniv.nl} \\ \makebox[.4\linewidth]{\texttt{leif.azzopardi@strath.ac.uk}} \makebox[.25\linewidth]{\texttt{shashemi@ebay.com}}
  \makebox[.25\linewidth]{\texttt{m.derijke@uva.nl}}
}

\usepackage{makecell}
\usepackage{svg}
\usepackage{ntheorem}
\theoremseparator{:}

\usepackage{xcolor,colortbl}
\usepackage{tabularx}
\usepackage{mathtools}
\usepackage{amsmath}
\usepackage{bbm}
\usepackage{amssymb}

\newcommand{\highlighter}{\cellcolor{gray!20}}
\newcommand{\lighthighghlighter}{\cellcolor{gray!10}}

\title{Evaluation of Attribution Bias in Generator-Aware\\Retrieval-Augmented Large Language Models}

\begin{document}

\maketitle

\begin{abstract}
Attributing answers to source documents is an approach used to enhance the verifiability of a model's output in retrieval-augmented generation (RAG).
Prior work has mainly focused on improving and evaluating the attribution quality of large language models (LLMs) in RAG, but this may come at the expense of inducing biases in the attribution of answers.
We define and examine two aspects in the evaluation of LLMs in RAG pipelines, namely attribution sensitivity and bias with respect to authorship information.
We explicitly inform an LLM about the authors of source documents, instruct it to attribute its answers, and analyze (i) how sensitive the LLM's output is to the author of source documents, and (ii) whether the LLM exhibits a bias towards human-written or AI-generated source documents. 
We design an experimental setup in which we use counterfactual evaluation to study three LLMs in terms of their attribution sensitivity and bias in RAG pipelines.
Our results show that 
adding authorship information to source documents 
can significantly change the attribution quality of LLMs by 3 to 18\%. We show that LLMs can have an attribution bias towards explicit human authorship, which can serve as a competing hypothesis for findings of prior work that shows that LLM-generated content may be preferred over human-written contents. Our findings indicate that metadata of source documents can influence LLMs' trust, and how they attribute their answers.
Furthermore, our research highlights attribution bias and sensitivity as a novel aspect of the brittleness of LLMs.
\end{abstract}
\section{Introduction}
\begin{figure}[t]
    \centering
    \includegraphics[width=\linewidth]{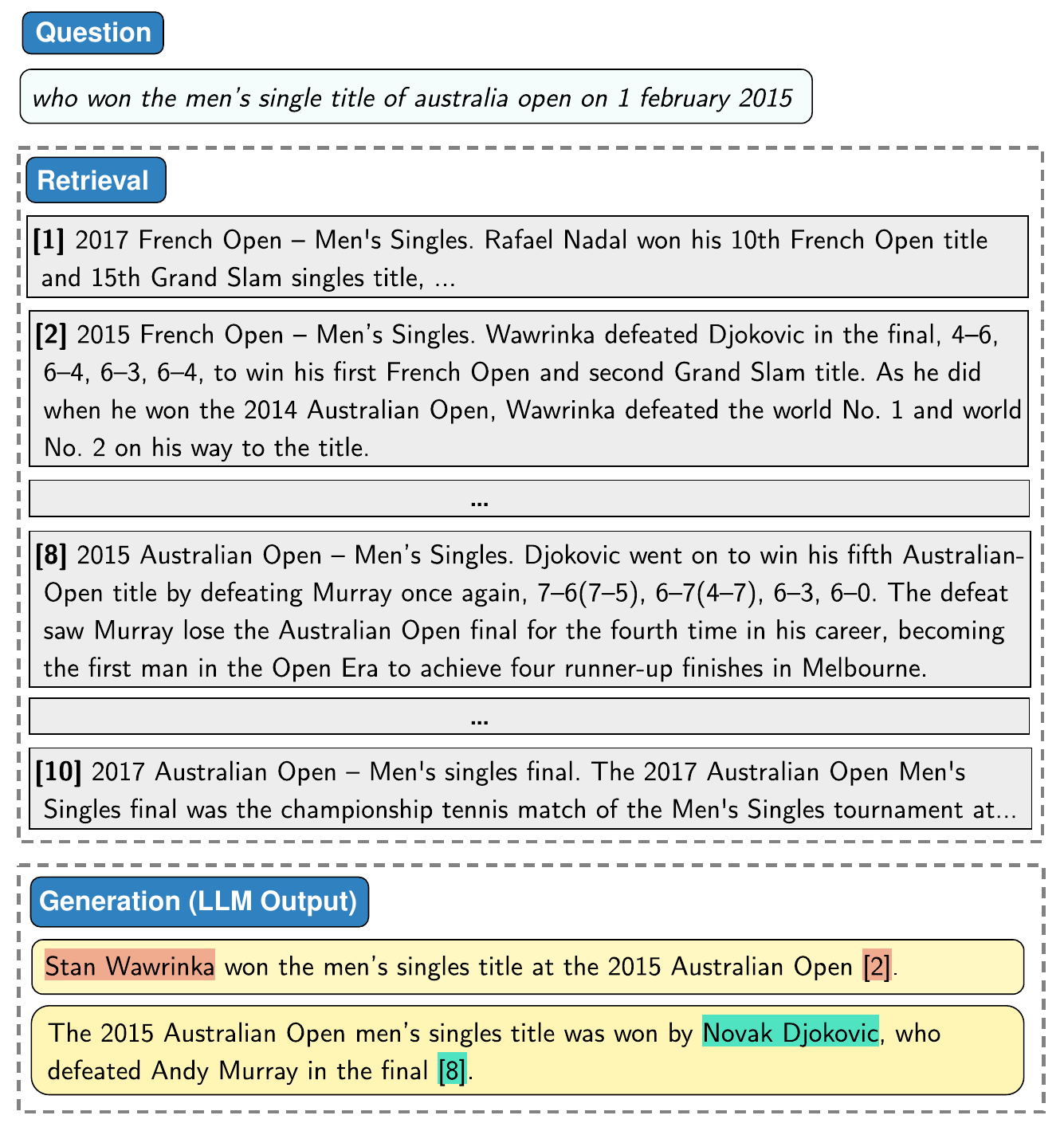}
    \caption{Retrieval-augmented answer/attribution generation using two LLMs. Together with the question, retrieval results are given to the LLMs in order to generate the answer.
    }
    \label{fig:topfigure}
\end{figure}
The goal of retrieval-augmented generation (RAG) is to generate an answer to a given question using a set of top-$k$ retrieved documents as context \cite{lewis2020retrieval}. Large language models (LLMs) have been a crucial part of RAG pipelines, mainly as the generator component \cite{jeong2024adaptive,li2024traq,lee2024planrag,asaiself}. Although the use of LLMs offers potential benefits, it also presents considerable risks, as they are prone to generate false or hallucinated claims \cite{ji2023survey}. This is important as such claims may misguide users, particularly when they are being used in critical fields such as the legal or medical domain \cite{augenstein2023factuality,malaviya2024expertqa,xiong2024medicinebenchmarking}.

Enabling LLMs to attribute their answer to the source of information has been proposed as a promising direction towards reducing the likelihood of such potential harms \cite{li2024towardsverifiable,patel2024towards,attributionbench2024li}. This attribution could assist users in tracing and understanding the basis of the information that LLMs are generating \cite{gao2023alce,huang2024citation}. There are many prior studies on answer attribution in RAG pipelines \cite{li2024towardsverifiable,bohnet2022attributed,muller2023evaluating,stolfo2024groundedness,hu2024benchmarking,menick2022teaching}. 

As Figure \ref{fig:topfigure} illustrates, LLMs are susceptible to making mistakes when attributing their answers to the input documents in RAG.
Moreover, enabling LLMs in RAG to attribute their answer may come at the expense of inducing biases, as LLMs may carry potential biases~\cite{ziems2024measuringbias,ozeki2024exploringbias,xie2023adaptive,esiobu2023robbie}. For instance, \citet{tan2024blinded} show that retrieval-augmented LLMs can be biased towards selecting their own generated text when this kind of content is present in their input.
Inspecting these biases is of paramount importance as they can be leveraged for both positive and negative purposes. 

In this paper, we study the performance of LLMs in terms of \emph{attribution sensitivity} and \emph{attribution bias} w.r.t. authorship information. When we explicitly inform LLMs about the authors of input documents, and instruct them to attribute their answers to the input documents (by providing citations to these documents), how sensitive are they to the authorship information of input documents? And are they biased towards either human or LLM authorship of input documents? To address these questions, we design a simulated evaluation setup in which we measure to what extent knowing the type of author of input documents affects the quality of attribution (citation) in LLMs. 

Prior work has indicated that LLM-generated content may consistently outperform human-authored content in search rankings, which, in turn, results in reducing the presence and exposure of human contributions online \citep{chen2024spiral,dai2024neural}. Inspired by these studies, we compare human-written documents against LLM-generated documents. 
We follow prior work in attribution generation by prompting LLMs to generate citations to the input documents \cite{gao2023alce,yue2023automatic}. Furthermore, we use counterfactual evaluation \cite{huang2020reducing,howard2024socialcounterfactuals,abolghasemi2024measuring, goldfarb2023bias,xie2023counter} to measure both authorship sensitivity and authorship bias of LLMs in RAG pipelines. 
This approach can be used more generally to measure algorithmic sensitivity or bias in a model or method: using counterfactual scenarios to see if changing certain characteristics
leads to different outcomes. 

Our experimental results show that three LLMs (\texttt{Mistral}, \texttt{Llama3} and \texttt{GPT-4}) are sensitive to authorship information that is included in the input documents prior to the generation. Moreover, we show that these models carry a bias towards human authorship against LLM authorship: they are more likely to attribute their answers to documents that are explicitly labelled as having been written by humans (even if the documents are actually generated by LLMs). 

We summarize our contributions as follows:
\vspace{-1mm}
\begin{itemize}[noitemsep,leftmargin=*]
    \item We define and study attribution sensitivity and bias w.r.t.\ authorship information, as a novel aspect of trustworthiness and brittleness in retrieval-augmented LLMs.

    \item We propose a systematic evaluation framework for measuring attribution sensitivity and bias.

    \item We show that adding authorship information (as metadata) to source documents may lead to statistically significant changes in the attribution quality of retrieval-augmented LLMs.

    \item We show that LLMs may have an attribution bias towards explicit human \emph{authorship}, which can serve as a competing hypothesis for findings of prior work
    that shows that LLM-generated \emph{content} is preferred over human-written \emph{content} by LLMs.\footnote{Our code is available at \url{https://github.com/aminvenv/attrieval}}
\end{itemize}

\section{Background}
\vspace{-2mm}
\textbf{Retrieval-Augmented Generation.}
Given a question $q$ and a set of top-$k$ retrieved documents $\mathcal{D}=$\{$d_1, d_2, \ldots, d_k$\} from a collection $\mathcal{C}$, the goal of retrieval-augmented generation (RAG) is to generate an answer for $q$ using $\mathcal{D}$ as context. LLMs are currently an important component of RAG pipelines, acting as the generator. The generator is given $q$, $\mathcal{D}$, and an instruction prompt on how to generate the answer \cite{jeong2024adaptive,li2024traq,lee2024planrag}. Using top-$k$ retrieved documents helps LLMs to be exposed to information that it might not have been trained/fine-tuned with during development. These documents are commonly retrieved using an effective sparse and/or dense retriever \cite{lewis2020retrieval,rau2024bergen}.   
\header{Attributive RAG}
    LLMs are prone to generate hallucinated (and even factually incorrect) answers \cite{rawte2023troubling,ji2023survey,yue2024evidence}. Attributing answers in RAG with LLMs is an approach taken as a step towards ensuring the veracity of the output of these models \citep{li2024towardsverifiable,bohnet2022attributed,hu2024benchmarking,khalifa2024source,kamalloo2023hagrid}. 
\citet{menick2022teaching} teach language models to support answers with verified quotes.  \citet{ye2024effective} propose a learning-based framework in which they fine-tune LLMs to generate citations, as opposed to prompting or relying on post-hoc attribution. \citet{stolfo2024groundedness} analyzes
whether every generated sentence in the output of LLMs is grounded in the retrieved documents or the LLM's pre-training data.

\section{Methodology}
\label{subsec:evaluationmethod}
We aim to measure the attribution sensitivity and bias of LLMs in RAG settings. 
We investigate to what extent the attribution quality of LLMs is affected by authorship information. To this end, we use counterfactual evaluation \cite{bottou2013counterfactual,wang2021robustness,gardner2020evaluating}. Counterfactual evaluation has been used across various natural language processing and information retrieval tasks \cite{abolghasemi2024cause, abolghasemi2024measuring,huang2020reducing,howard2024socialcounterfactuals,goldfarb2023bias}. 
This approach evaluates how a model's predictions change when a specific feature or set of features is altered while keeping everything else constant. 
In our case, the change is to add authorship information to the input documents of an LLM in a RAG setting. By doing so, we can evaluate the model's reliance on, bias towards, or sensitivity to that feature. 
To this end, we first generate answers with LLMs in a RAG setting using three RAG modes, as shown in Figure \ref{fig:ragmodes}. 

\subsection{RAG Modes}
\label{subsec:ragmodes}
Given a query $q$ and a set of top-$k$ retrieved documents $\mathcal{D}_q$ for $q$, we define three modes, based on authorship information of these documents that we provide to the answer generator LLM.

\header{Vanilla RAG}
In this mode, each document in $\mathcal{D}$ is given to the LLMs without information about who the authors are. This is the plain input format for input documents as shown in the input prompt for \emph{vanilla} answer/attribution generation in Figure \ref{tab:vanillaprompt}.

\header{Authorship Informed RAG}
In this mode, we inform the LLM about the actual author of each document. We denote the authorship of either an LLM or a human using \texttt{[LLM]} and \texttt{[Human]} tokens as shown by Figure \ref{fig:informed_prompt} in the Appendix.\footnote{\amin{In Section \ref{appendix:more-authorship-labels} in the Appendix, we study and provide results on replacing \texttt{[Human]} with a set of actual \{firstname, lastname\} as authors.}}

\header{Counterfactual-Authorship Informed RAG}
In this mode, we assign counterfactual authorship for each document.
If a document is written by a human, the counterfactual authorship of this document is \texttt{[LLM]}. In contrast, if a document is generated by an LLM, its counterfactual authorship is \texttt{[Human]}. By doing so, we can investigate to what extent being written by either human or LLM affects the attribution quality of LLM. The prompt used for this mode is the same as the one for Authorship Informed RAG mode.

\begin{figure}[t]
    \centering
    \includegraphics[width=\linewidth]{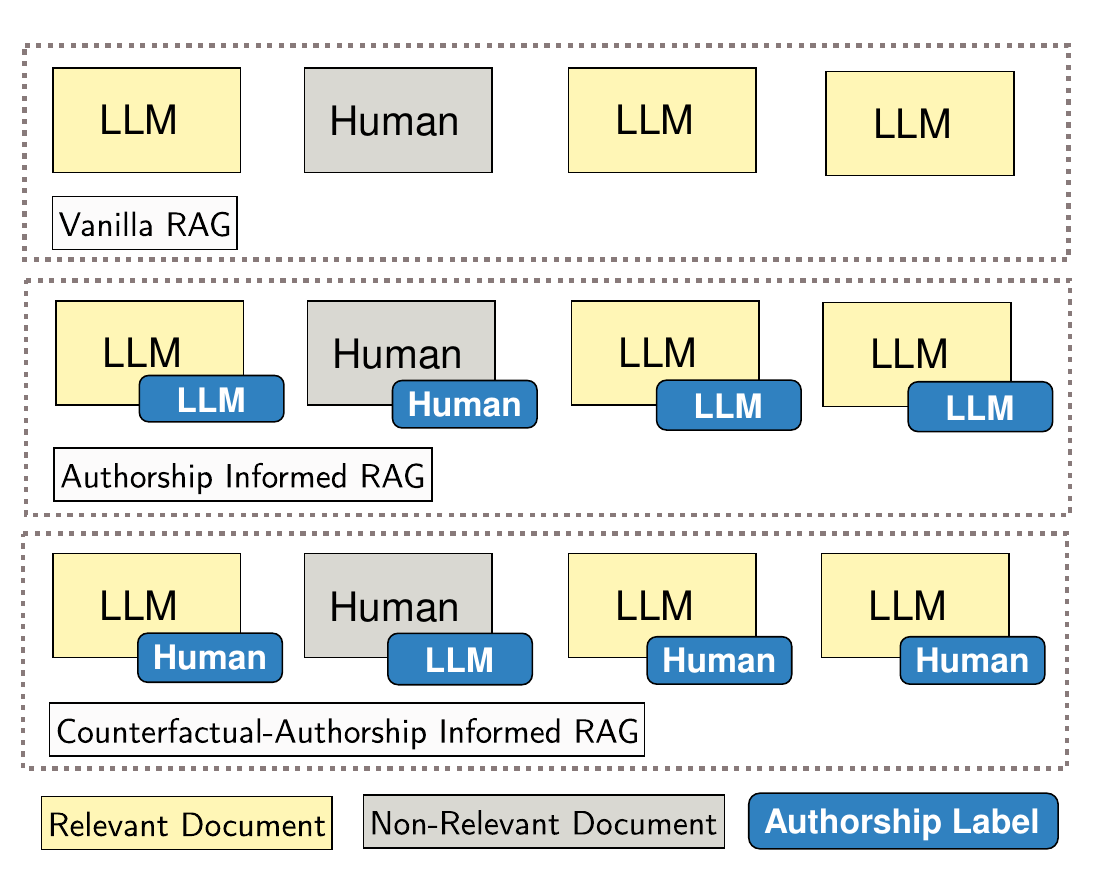}
    \caption{Three RAG modes (Section \ref{subsec:ragmodes}) for the setting with LLM actual authorship for relevant documents and Human actual authorship for non-relevant documents. The text in a rectangle denotes the actual generator (i.e., author) of each document. The text in the blue tags denotes the authorship label about which we inform the answer/attribution generator LLM.}
    \label{fig:ragmodes}
\end{figure}

Figure \ref{fig:ragmodes} shows the three RAG modes for a setting where the relevant documents are LLM-written and the non-relevant documents are human-written. 
\subsection{Answer/Attribution Generation}
In order to generate answers with each of the aforementioned RAG modes, we experiment with three LLMs: \texttt{Mistral}~\cite{jiang2023mistral}, \texttt{Llama3}~\cite{dubey2024llama} and \texttt{GPT-4}~\cite{openai2023gpt4}. Figure \ref{tab:vanillaprompt} shows the prompt used for \emph{vanilla} answer generation. Figure \ref{fig:informed_prompt} in the Appendix shows the prompt used for \emph{Authorship-Informed} and \emph{Counterfactual-Authorship Informed} answer generation. We follow prior work \cite{gao2023alce} in curating our prompts for this task.
\begin{table*}[ht]
    \centering
    \setlength{\tabcolsep}{4.5pt}
\begin{tcolorbox}
[boxrule=0.2mm, colback=gray!2, 
                  colframe= darkgray,
                  rounded corners, 
                  arc=0mm, 
                  ]
                  \fontsize{9}{11
                  }\selectfont
\ttfamily{\textbf{Instruction}: Write a concise answer for the given question (query) based on the provided search result documents, and cite them properly using [0][1][2] etc. 
\\ \\Please take these strict considerations into account during answer generation:
\\1. Documents are retrieved by a search engine. As such, not all the documents are relevant to the query. Only use and cite the relevant documents that contain the answer.
\\2. Do not analyze irrelevant documents.
\\ \\ \textbf{Search Results}:
\\ \\ Document [0](\{\textbf{\textcolor{orange}{text of Document [0]}}\})
\\  Document [1](\{\textbf{\textcolor{orange}{text of Document [1]}}\})
\\ ...
\\  Document [9](\{\textbf{\textcolor{orange}{text of Document [9]}}\})
\\ \\ \textbf{Question}: \{\textcolor{orange}{\textbf{query}}\}.
}
\end{tcolorbox}

    \captionof{figure}{Prompt used for vanilla retrieval-augmented answer generation. 
    }
    \label{tab:vanillaprompt}
\end{table*}

\subsection{Evaluation Metrics}
\textbf{Attribution Quality.}
We use precision and recall for evaluating the quality of attribution, i.e., how well the LLMs cite the relevant input documents. Precision of attribution for a single query is the fraction of correct citations among all cited documents in the output of an LLM. Recall is the fraction of cited relevant documents out of all relevant documents \cite{djeddal2024evaluation}. We use the queries that have only one relevant document containing the ground-truth answer in their top-$k$ retrieved list of documents.

\header{Attribution Sensitivity}
\label{subsubsec:method-attribution-sensitivity}
In order to measure the sensitivity of LLMs in RAG pipelines towards knowing authors of input documents in comparison to not knowing it, we use counterfactual evaluation and define a metric called Counterfactually-estimated Attribution Sensitivity (CAS):
\begin{align}
  \operatorname{CAS} (Q) =\frac{1}{\vert Q \vert } \sum_{q \in Q}^{}{\vert M_{\text{Informed}}^q- M_{\text{Vanilla}}^q} \vert.
  \label{equation:cas}
\end{align}
Here, $M^q$ represents the precision and recall metrics for query $q$, i.e., attribution quality for query $q$. For a single query $q$, CAS measures the difference between a base setup (the vanilla RAG mode) and a counterfactual setup (the authorship informed RAG mode) for the same set of input documents.
\header{Attribution Bias}
In order to measure the attribution bias of LLMs in RAG pipelines we use counterfactual evaluation and define a metric called Counterfactually-estimated Attribution Bias (CAB):
\begin{equation}
\mbox{}\hspace*{-1mm}
  \operatorname{CAB} (Q) = \frac{\omega}{\vert Q \vert} \sum_{q \in Q}^{}{M^q_{\text{Informed}} \!-\! M^q_{\text{CF-informed}}}
\hspace*{-1mm}\mbox{}
  \label{equation:cab}
\end{equation}
\begin{align}
\mbox{}\hspace*{-2mm}
  \omega \!=\!
    \begin{cases}
      \phantom{-}1,& \!\!\!\text{if } L_f(\mathcal{R})\!\!=\!\!\texttt{[Human]}, L_f(\mathcal{N})\!\!=\!\!\texttt{[LLM]}
      \hspace*{-3mm}\mbox{}\\ -1,& \!\!\!\text{otherwise.}
    \end{cases}
    \label{eq:omega}
\end{align}
Here, $M^q$ represents the precision and recall metrics, i.e., attribution quality, for query $q$, given the set of retrieved relevant documents $\mathcal{R}$, and the set of retrieved non-relevant documents $\mathcal{N}$. $L_f(\mathcal{X})$ stands for the authorship label of the set of documents $\mathcal{X}$ in the first term of Eq. \ref{equation:cab}, i.e., corresponding to $M^q_{\text{Informed}}$. For example, if we use human-written version of relevant documents ($\mathcal{R}$), and LLM-written version of non-relevant document ($\mathcal{N}$), and we label them with their actual generators (authors), i.e., we use authorship-informed RAG mode, then $L_f(\mathcal{R})$ is equal to \texttt{[Human]}, and $L_f(\mathcal{R})$ is equal to \texttt{[LLM]}. CAB measures the difference between metric values of a base setup (the Authorship Informed RAG mode) and a counterfactual setup (the Counterfactual-authorship Informed RAG mode) for the same set of input documents consisting of $\mathcal{R}$, and $\mathcal{N}$.
$\omega$ determines the direction of bias towards either human or LLMs: if the set of relevant documents ($\mathcal{R}$) and non-relevant documents ($\mathcal{N}$) are respectively written by \texttt{Human} and \texttt{LLM} (i.e., $L_f(\mathcal{R})$ = \texttt{[Human]}, $L_f(\mathcal{N})$ = \texttt{[LLM]}), for a single query, a positive difference $(M_{\text{Informed}}-M_{\text{CF-informed}})$ indicates bias towards human authorship, and a negative difference shows bias towards LLM authorship. In contrast, if the set of relevant documents ($\mathcal{R}$) and non-relevant documents ($\mathcal{N}$) are respectively written by \texttt{LLM} and \texttt{Human} (i.e., $L_f(\mathcal{R})$ = \texttt{[LLM]}, $L_f(\mathcal{N})$ = \texttt{[Human]}), a negative difference   $(M_{\text{Informed}}- M_{\text{CF-informed}})$ indicates a bias towards human authors, and a positive difference shows bias towards LLMs. We use $\omega$ to align these two conditions of actual authorship of input documents.

\header{Attribution Confidence}
To better explore the performance of LLMs in attribution generation, we analyze whether the LLMs are more confident when they attribute to certain types of document. To this aim, we look into the average probability of generation for attribution tokens, i.e., citation numbers (0, 1, \ldots):
\begin{equation}
    \mbox{}\hspace*{-2mm}
    \operatorname{AC}(\mathcal{S}) \!=\! 
    \frac{
    \sum_{q \in Q}\!
    \sum_{c_i \in C_{q}}^{}
    \!p(c_i| q, \mathcal{D}_q) \!\cdot\! \mathbbm{1}[c_i \!\in\! \mathcal{S}]
    }{
    \vert \sum_{q \in Q}^{}{\sum_{c_i \in C_{q}}^{}{\mathbbm{1}[c_i \in \mathcal{S}] \vert}
    }},
    \hspace*{-2mm}\mbox{}
    \label{equation:confidence}
\end{equation}
where $q$ is a query in the set of queries $Q$, $\mathcal{D}_q$ is the top-$k$ retrieved documents for $q$, $C_{q}$ stands for all attribution numbers in the answer to $q_i$, $c_i \in \{0,1,\ldots,k\}$, $\mathcal{S}$ indicates a set of documents, e.g., the set of relevant documents for all queries, and $p(c_i| q, \mathcal{D}_q)$ shows the probability of generation for the attribution token $c_i$ in the answer provided by LLM given query $q$ and its top-$k$ retrieved documents $\mathcal{D}_q$. $\mathbbm{1}[c_i \in \mathcal{S}]$ equals 1 if $c_i \in \mathcal{S}$.

\header{Answer Correctness} In order to evaluate the quality of the generated answer, we follow \cite{petroni2021kilt,gao2023alce} and use automatic evaluation. Following \cite{gao2023alce,stolfo2024groundedness}, we use the normalized human-generated answer in the benchmark as the ground-truth answer and adopt Exact Match (EM) \cite{siriwardhana2023improving,wang2023learning} as the evaluation metric for answer correctness (see example in Figure \ref{tab:example_dragon}).

\section{Experimental Settings}
\label{sec:experimental-settings}
\textbf{Models.} We use \texttt{gpt-4-0314}~\cite{openai2023gpt4}, \texttt{meta-llama/Meta-Llama-3-8B-Instruct},\footnote{\url{https://huggingface.co/meta-llama/Meta-Llama-3-8B-Instruct}} and  \texttt{mistralai/Mistral-7B-Instruct-v0.3}\footnote{\url{https://huggingface.co/mistralai/Mistral-7B-Instruct-v0.3}} as answer generator LLMs in our RAG pipelines. We refer to these models as \texttt{GPT-4}, \texttt{Llama3}, and \texttt{Mistral}, respectively.
\header{Benchmarks} We use two benchmarks in our experiments: Natural Questions (NQ) \cite{kwiatkowski2019natural} and MS MARCO Question Answering \cite{bajaj2016ms} (to which we refer as MS MARCO). For each benchmark, we randomly sample 500 queries. To retrieve top-$k$ passages for each query in the NQ benchmark, we use BM25, a widely-used lexical matching retrieval model.
For queries in the MS MARCO benchmark, we use passages that are extracted from relevant web documents using the state-of-the-art passage retrieval system at Bing \cite{bajaj2016ms}. We note that we study the effect of different retrievers and different number of retrieved source documents in Section \ref{app:topk-effect} and \ref{app:retrievers} in the Appendix, respectively.

\header{Synthetic Collection}
To construct LLM-written documents, we use \texttt{Llama3} to re-write a given document from our collections without adding/removing information to/from the document. Specifically, we use a low temperature close to 0 as it makes the LLM extremely restrictive, focusing only on the most probable tokens resulting in (highly) deterministic outputs. The reason for not generating the documents from scratch is to make sure we keep the relevance/non-relevance status of documents w.r.t a query.
To ensure the quality of synthetic passages, we conduct a number of annotation steps using two expert annotators. This is detailed in Section \ref{appendix:data-quality} in the Appendix.
Importantly, in Section \ref{sec:mixed-rag-mode}, we show that even without using LLM-generated documents (i.e., only designating \texttt{[Human]} and \texttt{[LLM]} as authors of documents to the original input documents) findings are the same as when we use actual LLM-generated documents.

\section{Experimental Results}
In this section, we explore the performance of LLMs for attributing their answer to top-$k$ retrieved source documents using the evaluation metrics introduced in Section~\ref{subsec:evaluationmethod}. All significance tests in the result tables are according to a paired t-test with $p<0.05$. 
\begin{table*}[ht]
    \centering
        \renewcommand{\arraystretch}{1}
    \scalebox{0.85
    }{
    \begin{tabular}{l ll l ccc }
        \toprule

        \multirow{2}{*}{\makecell{Answer \\ generator}} & \multirow{2}{*}{\makecell{Relevant \\ documents}} & \multirow{2}{*}{\makecell{Non-relevant \\ documents}}
        & \multirow{2}{*}{\makecell{RAG \\ mode}} &  \multicolumn{2}{c}{Attribution quality} & Correctness 
        \\ \cmidrule(r){5-6} \cmidrule(r){7-7}  
        &  & 
        &  & Precision &  Recall  & EM

      \\ \midrule
       \multicolumn{1}{l}{\lighthighghlighter \footnotesize \texttt{NQ}} & \lighthighghlighter &  \lighthighghlighter & \lighthighghlighter & \lighthighghlighter & \lighthighghlighter & \lighthighghlighter 
        \\
      \multirow{6}{*}{Mistral} & \multirow{3}{*}{LLM} & \multirow{3}{*}{Human} 
         &    Vanilla & 47.6 & 76.6  &  0.722
         \\ 
         & &
         &  \highlighter Informed 
          & \highlighter 42.1  & \highlighter 68.2  & \highlighter 0.730 
        \\ 
        & & 
        & 
           CF-informed
        &    \textbf{52.7}\rlap{$^{\dagger}$} &    \textbf{77.8}\rlap{$^{\dagger}$}  &      \textbf{0.738}
         
         \\ \cmidrule(r){2-7}
         & \multirow{3}{*}{Human} & \multirow{3}{*}{LLM} 
         &    Vanilla  & 51.0 & 78.4  & \textbf{0.776}  
         \\ 
         & &
         & \highlighter Informed 
          &  \highlighter \textbf{53.4}\rlap{$^{\dagger}$} &  \highlighter \textbf{77.8}\rlap{$^{\dagger}$} &  \highlighter 0.774
         \\
         & & 
         &    CF-informed 
         & 44.0 & 70.2 & 0.772
         
         \\ \midrule
         \multirow{6}{*}{Llama3} & \multirow{3}{*}{LLM} & \multirow{3}{*}{Human} 
         &    Vanilla &    49.2 &     69.2 &   0.742
         \\ 
         & &
         &  \highlighter Informed 
         &  \highlighter 45.4  & \highlighter  69.6   & \highlighter 0.730
        \\ 
        & & 
        & 
           CF-informed
        &    \textbf{57.2}\rlap{$^{\dagger}$} &    \textbf{77.6}\rlap{$^{\dagger}$}  &      \textbf{0.748}
         
         \\ \cmidrule(r){2-7}
         & \multirow{3}{*}{Human} & \multirow{3}{*}{LLM} 
         &    Vanilla &    53.5 &    71.0 &   0.766
         \\ 
         & &
         & \highlighter Informed 
          &  \highlighter \textbf{59.9}\rlap{$^{\dagger}$} &  \highlighter \textbf{77.8}\rlap{$^{\dagger}$} &  \highlighter \textbf{0.790}
         \\
         & & 
         &    CF-informed 
         &    44.8
 &    69.2
  &    0.762

         \\ \cmidrule(r){1-7}

         \multirow{6}{*}{GPT-4} &  \multirow{3}{*}{LLM} & \multirow{3}{*}{Human} 
         &    Vanilla &    63.3  &    68.8  &      0.736
         \\ 
         & &
         & \highlighter Informed 
         &  \highlighter 59.7  & \highlighter 64.6 &  \highlighter 0.740
         \\ 
         & &
         &    CF-informed
         &    \textbf{65.9}\rlap{$^{\dagger}$}  &    \textbf{72.2}\rlap{$^{\dagger}$}&        \textbf{0.742}
         
         \\ \cmidrule(r){2-7}
         & \multirow{3}{*}{Human} & \multirow{3}{*}{LLM} 
         &  Vanilla &  64.1 &  68.8 &  0.760
         \\ 
         &  &
         &  \highlighter  Informed 
         &   \highlighter  \textbf{66.1} &  \highlighter  \textbf{72.2}\rlap{$^{\dagger}$} &     \highlighter \textbf{0.776} 
         \\ 
         &  & 
         &  CF-informed
         &  60.3  &  65.0 &  0.758

         \\ \bottomrule
    \end{tabular}
    
    }
    \caption{Quality of attribution and answer correctness. The columns ``Relevant Documents'' and ``Non-relevant Documents'' refer to the actual authorship of input documents. Informed refers to the authorship-informed RAG and CF-informed refers to counterfactual-authorship informed RAG (Section \ref{subsec:ragmodes}). $\dagger$ indicates statistically significant improvements over the two other RAG Modes in each combination of ``Relevant'' and ``Non-relevant'' documents.
    }    
    \label{tab:groundness_table}
\end{table*}
\header{Attribution Quality}
Table \ref{tab:groundness_table} shows the results of attribution by three LLMs, \texttt{Mistral}, \texttt{Llama3} and \texttt{GPT-4}, under different settings for NQ benchmark. Besides, Table \ref{tab:attributionquality-msmarco} in the Appendix shows the same set of results for the MS MARCO benchmark. The two columns ``Relevant documents'' and ``Non-relevant documents'' indicate the actual generator (author) of these documents. The column ``RAG mode'' indicates how we inform the answer generator LLMs about the authorship label of relevant and non-relevant documents, as described in Section~\ref{subsec:ragmodes}: in the ``Vanilla'' RAG mode, no information regarding the generator (author) of the input source documents is given to the LLM. In the ``Informed'' RAG mode the LLM is informed about the actual generator of the input source documents, and in the ``CF-Informed'' RAG mode the LLM is provided with counterfactual authorship information. As Table \ref{tab:groundness_table} shows, the three LLMs (\texttt{Mistral}, \texttt{Llama3} and \texttt{GPT-4}) fall short of perfectly grounding their answers to the relevant documents of a given question, which is in line with the findings of prior work \cite{gao2023alce,djeddal2024evaluation,attributionbench2024li}. 
\begin{table}[ht]
    \centering  
    \setlength{\tabcolsep}{1.5pt}
    \renewcommand{\arraystretch}{1.1}

    \scalebox{0.84}{
        \begin{tabular}{l ll ccc }
        \toprule

        \multirow{2}{*}{\makecell{Answer \\ generator}} & \multirow{2}{*}{\makecell{Relevant \\ documents}} & \multirow{2}{*}{\makecell{Non-relevant \\ documents}}
        & \multicolumn{2}{c}{CAS}
        \\ \cmidrule(r){4-5} 
        &  & 
        & $\Delta$Precision & $\Delta$Recall 
        \\ \midrule
        \multicolumn{1}{l}{\lighthighghlighter \footnotesize \texttt{NQ}} & \lighthighghlighter &  \lighthighghlighter & \lighthighghlighter & \lighthighghlighter  
        \\

       \multirow{2}{*}{Mistral}  & \multirow{1}{*}{LLM} & \multirow{1}{*}{Human} 

         &  16.2\rlap{$^{\dagger}$} &  17.2\rlap{$^{\dagger}$}
         
         \\ 
         & \multirow{1}{*}{Human} & \multirow{1}{*}{LLM} 

         &  20.1   & 17.0
         
         \\ \midrule  
        \multirow{2}{*}{Llama3}  & \multirow{1}{*}{LLM} & \multirow{1}{*}{Human} 
         &  13.2\rlap{$^{\dagger}$} & 14.8 

         \\ 
         & \multirow{1}{*}{Human} & \multirow{1}{*}{LLM} 
         &  17.7\rlap{$^{\dagger}$}  &  16.0\rlap{$^{\dagger}$}

         \\ \midrule
          \multirow{2}{*}{GPT-4} &  \multirow{1}{*}{LLM} & \multirow{1}{*}{Human} 
         &  \phantom{0}9.7\rlap{$^{\dagger}$}  & 10.2\rlap{$^{\dagger}$}

         \\ 
         & \multirow{1}{*}{Human} & \multirow{1}{*}{LLM} 
         & \phantom{0}8.7  & \phantom{0}9.0\rlap{$^{\dagger}$} 

      \\ \midrule
       \multicolumn{1}{l}{\lighthighghlighter \footnotesize \texttt{MS MARCO}} & \lighthighghlighter &  \lighthighghlighter & \lighthighghlighter & \lighthighghlighter & \lighthighghlighter
        \\
       \multirow{2}{*}{Mistral}  & \multirow{1}{*}{LLM} & \multirow{1}{*}{Human} 

         &  10.9 &  21.4\rlap{$^{\dagger}$}
         
         \\ 
         & \multirow{1}{*}{Human} & \multirow{1}{*}{LLM} 

         &  12.9\rlap{$^{\dagger}$}   & 16.6
         
         \\ \midrule  
         \multirow{2}{*}{Llama3} & \multirow{1}{*}{LLM} & \multirow{1}{*}{Human} 
         &  12.9\rlap{$^{\dagger}$}  & 20.4\rlap{$^{\dagger}$} 

         \\ 
         & \multirow{1}{*}{Human} & \multirow{1
         }{*}{LLM} 
         &  17.8\rlap{$^{\dagger}$}   &  19.6\rlap{$^{\dagger}$} 

         \\ \cmidrule(r){1-5}

         \multirow{2}{*}{GPT-4} &  \multirow{1}{*}{LLM} & \multirow{1}{*}{Human} 
         &  \phantom{0}8.2\rlap{$^{\dagger}$}  &  \phantom{0}9.6\rlap{$^{\dagger}$}

         \\ 
         & \multirow{1}{*}{Human} & \multirow{1}{*}{LLM} 
         &  10.9  & 15.8\rlap{$^{\dagger}$}

         \\ \bottomrule
    \end{tabular}
    }
    \caption{Attribution sensitivity (CAS) results. Values range from 0 (minimum sensitivity) to 100 (maximum sensitivity). ${\dagger}$ indicates statistically significant values.}
    \label{tab:only_sensitivity}
\end{table}
\begin{table}[ht]
    \centering
    \setlength{\tabcolsep}{2pt}
    \renewcommand{\arraystretch}{1.1}
    
    \scalebox{0.84}{
        \begin{tabular}{l ll cc}
        \toprule

        \multirow{2}{*}{\makecell{Answer \\ generator}} & \multirow{2}{*}{\makecell{Relevant \\ documents}} & \multirow{2}{*}{\makecell{Non-relevant \\ documents}}
         & \multicolumn{2}{c}{CAB} 
        \\ \cmidrule(r){4-5}
        &  & 
        & $\Delta$Precision & $\Delta$Recall

       \\ \midrule
       \multicolumn{1}{l}{\lighthighghlighter \footnotesize \texttt{NQ}} & \lighthighghlighter &  \lighthighghlighter & \lighthighghlighter & \lighthighghlighter
        \\
       \multirow{2}{*}{Mistral}  & \multirow{1}{*}{LLM} & \multirow{1}{*}{Human} 

         & +10.6\rlap{$^{\dagger}$}   &  \phantom{0}+9.6\rlap{$^{\dagger}$} 
         
         \\ 
         & \multirow{1}{*}{Human} & \multirow{1}{*}{LLM} 

         &  \phantom{0}+9.4\rlap{$^{\dagger}$}    & \phantom{0}+7.6\rlap{$^{\dagger}$}
         
         \\ \midrule 
         \multirow{2}{*}{Llama3} & \multirow{1}{*}{LLM} & \multirow{1}{*}{Human} 

         &  +11.8\rlap{$^{\dagger}$}   &   \phantom{0}+8.0\rlap{$^{\dagger}$}
         \\ 
         & \multirow{1}{*}{Human} & \multirow{1
         }{*}{LLM} 

         & +15.1\rlap{$^{\dagger}$} & \phantom{0}+8.6\rlap{$^{\dagger}$}
         \\ \cmidrule(r){1-5}

         \multirow{2}{*}{GPT-4} &  \multirow{1}{*}{LLM} & \multirow{1}{*}{Human} 

         &   \phantom{0}+6.2\rlap{$^{\dagger}$}  &  \phantom{0}+7.6\rlap{$^{\dagger}$}

         \\ 
         & \multirow{1}{*}{Human} & \multirow{1}{*}{LLM} 

         &  \phantom{0}+5.8\rlap{$^{\dagger}$}   & \phantom{0}+7.2\rlap{$^{\dagger}$}

        \\ \midrule
        \multicolumn{1}{l}{\lighthighghlighter \footnotesize \texttt{MS MARCO}} & \lighthighghlighter &  \lighthighghlighter & \lighthighghlighter & \lighthighghlighter
        \\
       \multirow{2}{*}{Mistral}  & \multirow{1}{*}{LLM} & \multirow{1}{*}{Human} 

         & \phantom{0}+9.5\rlap{$^{\dagger}$}  &  +13.8\rlap{$^{\dagger}$} 
         
         \\ 
         & \multirow{1}{*}{Human} & \multirow{1}{*}{LLM} 

         &  \phantom{0}+8.0\rlap{$^{\dagger}$}   & +12.4\rlap{$^{\dagger}$} 
         
         \\ \midrule 
        \multirow{2}{*}{Llama3}  & \multirow{1}{*}{LLM} & \multirow{1}{*}{Human} 

         &  +15.6\rlap{$^{\dagger}$}   &  +18.2\rlap{$^{\dagger}$} 
         
         \\ 
         & \multirow{1}{*}{Human} & \multirow{1}{*}{LLM} 

         &   +15.1\rlap{$^{\dagger}$}  & +16.4\rlap{$^{\dagger}$}
         
         \\ \midrule
          \multirow{2}{*}{GPT-4} &  \multirow{1}{*}{LLM} & \multirow{1}{*}{Human} 

         &   \phantom{0}+6.1\rlap{$^{\dagger}$}  & \phantom{0}+9.0\rlap{$^{\dagger}$}

         \\ 
         & \multirow{1}{*}{Human} & \multirow{1}{*}{LLM} 

         &  \phantom{0}+5.4\rlap{$^{\dagger}$}  & +10.8\rlap{$^{\dagger}$} 

         \\ \bottomrule
    \end{tabular}
    
    }
    \caption{Attribution Bias (CAB) results.
    Values range from -100 (completely biased towards LLM authorship) to +100 (completely biased towards human authorship). $\dagger$ indicates statistically significant bias values.
    }
    \label{tab:only_bias}
\end{table}
\header{Attribution Sensitivity and Bias}
Table \ref{tab:only_bias} shows the attribution bias results in terms of CAB (Eq.~\ref{equation:cab}). All three LLMs, \texttt{Mistral}, \texttt{Llama3}, and \texttt{GPT-4}, carry a bias towards human authorship in the input documents. 
Moreover, on both datasets, NQ and MS~MARCO, \texttt{Mistral} and \texttt{Llama3} have higher bias values than \texttt{GPT-4}. Besides, Table \ref{tab:only_sensitivity} shows the attribution sensitivity results in terms of CAS (Eq.~\ref{equation:cas}). All three LLMs, \texttt{Mistral}, \texttt{Llama3}, and \texttt{GPT-4}, show sensitivity towards the inclusion of authorship information for the input documents of LLMs. Moreover, similar to the attribution bias values in Table \ref{tab:only_bias}, \texttt{Mistral} and  \texttt{Llama3} carry a higher attribution sensitivity than \texttt{GPT-4}. We note that we conducted experiments using different prompts and observed that the findings remained consistent across multiple runs.

\begin{figure}[ht]
    \centering   \includegraphics[width=\linewidth]{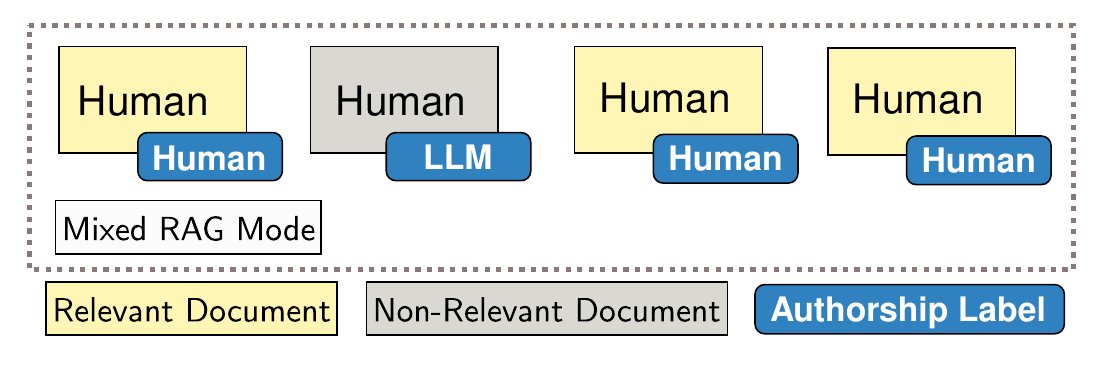}
    \caption{Mixed RAG mode for the setting where we use original human-authored documents. In this example, we have ``Informed'' mode for relevant documents and ``CF-Informed'' for non-relevant documents.}
    \label{fig:mix_mode}
\end{figure}

\header{Mixed RAG Mode}
\label{sec:mixed-rag-mode}
To better disentangle the effect of LLM generated text qualities (e.g., a potential implicit bias of LLMs towards LLM-written documents \cite{tan2024blinded}) from the impact of authorship information, we now use the same set of documents in the input of LLM in the RAG, and only use different authorship labels for relevant and non-relevant documents. For this new setup, to which we refer as the Mixed RAG mode, we evaluate both a complete set of synthetic documents (i.e., for both relevant and non-relevant) and a complete set of human-written documents. Figure \ref{fig:mix_mode} shows an example of Mixed RAG mode for the setting where we have human-written documents, with different authorship labels for relevant and non-relevant documents.
The CAB (Eq.~\ref{equation:cab}) for Mixed RAG mode is reformulated as follows:
\begin{equation}
\begin{split}
  \operatorname{CAB} (Q) = \frac{\omega}{\vert Q \vert} \sum_{q \in Q}^{} & M^q_{\text{Informed/CF-Informed}}-\\[-4pt] 
  & \mbox{}\hspace*{5mm} M^q_{\text{CF-Informed/Informed}},
\end{split}  
  \label{equation:cab_mixed}
\end{equation}
where $X$ and $Y$ in $M^q_{X/Y}$ stand for the RAG mode for the set of relevant documents and the set of non-relevant documents, respectively.
The results of attribution quality for Mixed-RAG modes are shown in Table \ref{tab:brief_mixed_NQ}.\footnote{See Tables \ref{tab:mixed_full1} and  \ref{tab:mixed_full2} (Appendix) for the complete set of results.} We see that, similar to Table \ref{tab:groundness_table}, across different settings, when the relevant documents are labeled with human-authorship and non-relevant ones are labeled with LLM-authorship, the attribution quality is higher than the other way around.%
\begin{table*}[ht]
    \centering
    \scalebox{0.85}{
    \setlength{\tabcolsep}{2.8pt}
      \begin{tabular}{l ll ll ccc }
        \toprule

        \multirow{2}{*}{\makecell{Answer \\ generator}} & \multirow{2}{*}{\makecell{Relevant \\ documents}} & \multirow{2}{*}{\makecell{Non-relevant \\ documents}}
        & \multicolumn{2}{c}{Mixed RAG mode} &  \multicolumn{2}{c}{Attribution quality} & Correctness 
        \\ \cmidrule(r){4-5} \cmidrule(r){6-7} \cmidrule(r){8-8} 
        &  & & Relevant
        & Non-relevant  & Precision &  Recall & EM
        \\ \midrule
        \multicolumn{1}{l}{\lighthighghlighter \footnotesize \texttt{NQ}} & \lighthighghlighter &  \lighthighghlighter & \lighthighghlighter & \lighthighghlighter & \lighthighghlighter & \lighthighghlighter &
        \lighthighghlighter
        \\
   \multirow{4}{*}{Mistral}& \multirow{2}{*}{Human} & \multirow{2}{*}{Human} & 
        

         CF-informed &  Informed &  44.8 &  71.8 &  0.772

        \\
        &&& 
        Informed & CF-informed & \textbf{52.3}\rlap{$^{\dagger}$} & \textbf{77.2}\rlap{$^{\dagger}$}&  \textbf{0.780}

        \\ \cmidrule(r){2-8}
        & \multirow{2}{*}{LLM} & \multirow{2}{*}{LLM} & 
        
         CF-informed &  Informed &  \textbf{48.7}\rlap{$^{\dagger}$} &  \textbf{74.6}\rlap{$^{\dagger}$}
        &   0.718
        \\
        &&& 
        Informed & CF-informed &  42.9
 &  69.4
 & \textbf{0.742}
       \\ \midrule
        \multirow{4}{*}{Llama3}& \multirow{2}{*}{Human} & \multirow{2}{*}{Human} &

         CF-informed &  Informed &  45.7 &  69.6 &  0.784    

        \\
        &&& 
        Informed & CF-informed & \textbf{57.4}\rlap{$^{\dagger}$} & \textbf{77.6}\rlap{$^{\dagger}$}&  \textbf{0.808}

        \\ \cmidrule(r){2-8}
        & \multirow{2}{*}{LLM} & \multirow{2}{*}{LLM} &

         CF-informed &  Informed &  \textbf{59.3}\rlap{$^{\dagger}$} &  \textbf{77.8}\rlap{$^{\dagger}$} 
        &   \textbf{0.744}
        \\
        &&& 
        Informed & CF-informed & 44.7
 & 68.4
 & 0.726

        \\ \midrule
        \multirow{4}{*}{GPT-4}& \multirow{2}{*}{Human} & \multirow{2}{*}{Human} &

         CF-informed &  Informed &  65.8 &   70.6 &  \textbf{0.794}
        \\
        & & & 
        Informed & CF-informed & \textbf{69.1}\rlap{$^{\dagger}$} &  \textbf{74.0}\rlap{$^{\dagger}$}  & 0.784

        \\ \cmidrule(r){2-8}
        & \multirow{2}{*}{LLM} & \multirow{2}{*}{LLM} &

         CF-informed &  Informed & \textbf{66.1}\rlap{$^{\dagger}$} &  \textbf{71.2}\rlap{$^{\dagger}$}  &   \textbf{0.730}
        \\
        &&& 
        Informed & CF-informed & 61.7 &  66.8 & 0.716

         \\ \bottomrule
    \end{tabular}
    
    }
    \caption{Quality of attribution and answer correctness for Mixed RAG mode. The columns ``Relevant Documents'' and ``Non-relevant Documents'' refer to the actual authorship of input documents. $\dagger$ indicates statistically significant improvements over the other Mixed RAG mode in each combination of relevant and non-relevant documents.
    }
    \label{tab:brief_mixed_NQ}
\end{table*}
\begin{table}[t]
    \centering
    \setlength{\tabcolsep}{2pt}
    \renewcommand{\arraystretch}{1}
    \scalebox{0.82}{
        \begin{tabular}{l ll cc }
        \toprule

        \multirow{2}{*}{\makecell{Answer \\ generator}} & \multirow{2}{*}{\makecell{Relevant \\ documents}} & \multirow{2}{*}{\makecell{Non-relevant \\ documents}}
         & \multicolumn{2}{c}{CAB} 
        \\ \cmidrule(r){4-5}
        &  & 
        & $\Delta$Precision & $\Delta$Recall

             \\ \midrule
       \multicolumn{1}{l}{\lighthighghlighter \footnotesize \texttt{NQ}} & \lighthighghlighter &  \lighthighghlighter & \lighthighghlighter & \lighthighghlighter
        \\

       \multirow{2}{*}{Mistral}  & \multirow{1}{*}{Human} & \multirow{1}{*}{Human} 

         & \phantom{0}+7.5\rlap{$^{\dagger}$}  &  \phantom{0}+5.4\rlap{$^{\dagger}$}
         
         \\ 
         & \multirow{1}{*}{LLM} & \multirow{1}{*}{LLM} 

         &  \phantom{0}+5.8\rlap{$^{\dagger}$}   & \phantom{0}+5.2\rlap{$^{\dagger}$}
         
         \\ \midrule  
         
         \multirow{2}{*}{Llama3} & \multirow{1}{*}{Human} & \multirow{1}{*}{Human} 

         &  +11.7\rlap{$^{\dagger}$}   & \phantom{0}+8.0\rlap{$^{\dagger}$} 
         \\ 
         & \multirow{1}{*}{LLM} & \multirow{1
         }{*}{LLM} 

         & +14.6\rlap{$^{\dagger}$}  & \phantom{0}+9.4\rlap{$^{\dagger}$}
         \\ \cmidrule(r){1-5}

         \multirow{2}{*}{GPT-4} &  \multirow{1}{*}{Human} & \multirow{1}{*}{Human} 

         &  \phantom{0}+3.3\rlap{$^{\dagger}$}  & \phantom{0}+3.4\rlap{$^{\dagger}$}

         \\ 
         & \multirow{1}{*}{LLM} & \multirow{1}{*}{LLM} 

         &  \phantom{0}+4.4\rlap{$^{\dagger}$}  & \phantom{0}+4.4\rlap{$^{\dagger}$}

        \\ \midrule
        \multicolumn{1}{l}{\lighthighghlighter \footnotesize \texttt{MS MARCO}} & \lighthighghlighter &  \lighthighghlighter & \lighthighghlighter & \lighthighghlighter
        \\

       \multirow{2}{*}{Mistral}  & \multirow{1}{*}{Human} & \multirow{1}{*}{Human} 

         & \phantom{0}+8.6\rlap{$^{\dagger}$}  &  +14.8\rlap{$^{\dagger}$}
         
         \\ 
         & \multirow{1}{*}{LLM} & \multirow{1}{*}{LLM} 

         & \phantom{0}+8.7\rlap{$^{\dagger}$}    & +13.8\rlap{$^{\dagger}$}
         
         \\ \midrule  
         
        \multirow{2}{*}{Llama3}  & \multirow{1}{*}{Human} & \multirow{1}{*}{Human} 

         &  +12.6\rlap{$^{\dagger}$}   &  +10.4\rlap{$^{\dagger}$}
         
         \\ 
         & \multirow{1}{*}{LLM} & \multirow{1}{*}{LLM} 

         &  \phantom{0}+9.7\rlap{$^{\dagger}$}   & \phantom{0}+9.8\rlap{$^{\dagger}$}
         
         \\ \midrule
          \multirow{2}{*}{GPT-4} &  \multirow{1}{*}{Human} & \multirow{1}{*}{Human} 

         & \phantom{0}+7.4\rlap{$^{\dagger}$}    & \phantom{0}+9.4\rlap{$^{\dagger}$}

         \\ 
         & \multirow{1}{*}{LLM} & \multirow{1}{*}{LLM} 

         & \phantom{0}+5.4\rlap{$^{\dagger}$}  & \phantom{0}+5.2\rlap{$^{\dagger}$}
  
         \\ \bottomrule
    \end{tabular}
    
    }
    \caption{Attribution Bias (CAB) results for Mixed RAG modes. Positive values indicate a bias towards human. $\dagger$ indicates statistically significant bias values. Values range from -100 (completely biased towards LLM authorship) to +100 (completely biased towards human authorship).
    }
    \label{table_mixed_only_bias}
\end{table}
Moreover, Table \ref{table_mixed_only_bias} illustrates the attribution bias for Mixed RAG modes.
Similar to the results in Table \ref{tab:only_bias}, there is a bias towards human authorship in all three LLMs. This indicates the existence of authorship bias regardless of the origin of the input documents, i.e., the actual author of the input documents. Furthermore,  similar to the results in Table \ref{tab:only_bias}, \texttt{Mistral} and \texttt{Llama3} show higher rates of attribution bias than \texttt{GPT-4}. Additionally, we find that when we have the same authorship label on both relevant and non-relevant documents (rows with the same RAG mode for relevant and non-relevant documents in Tables \ref{tab:mixed_full1} and \ref{tab:mixed_full2} in the Appendix), we do not see consistent patterns as to how LLMs attribute the answers to the input documents. Finally, we note that in Section \ref{appendix:more-authorship-labels} of the Appendix, we show additional results using real-world names as authors which further indicates the presence of attribution bias and sensitivity in LLMs towards authorship information. 
\header{Attribution Confidence}
Using Eq.~\ref{equation:confidence}, we compute the confidence of LLMs when they attribute their answer to an input document. Table \ref{tab:confidence_table_NQ} shows the attribution confidence of LLMs for relevant and non-relevant documents.\footnote{Table \ref{tab:confidence_table_NLG} in the Appendix shows the results on MS MARCO.} 
Across the majority of RAG modes over different origins for relevant and non-relevant documents, the confidence of all three LLMs for attributing to relevant documents is higher than for attributing to non-relevant ones. 
We can also see that authorship labels (i.e., using different RAG modes) do not affect this outcome. In other words, it is being relevant or not that makes the difference here. These results indicate a promising direction for improving attribution in LLMs: low confidence of LLMs in attributing to a specific document might be a useful signal for the relevance of that document to a given query. 
\begin{table}[t]
    \centering
    \setlength{\tabcolsep}{1.2pt}
    \renewcommand{\arraystretch}{0.8}
    \scalebox{0.8}
    {%
        \begin{tabular}{l ll l cc }
        \toprule

        \multirow{2}{*}{\makecell{Answer \\ generator}} & \multirow{2}{*}{\makecell{Rel. \\ Docs.}} & \multirow{2}{*}{\makecell{Non-rel. \\ docs.}}
        & \multirow{2}{*}{\makecell{RAG \\ mode}} &  \multicolumn{2}{c}{Confidence (AC)} 
        \\ \cmidrule(r){5-6} 
        &  & 
        &  & Rel. &  Non-rel. 
      \\ \midrule
       \multicolumn{1}{l}{\lighthighghlighter \footnotesize \texttt{NQ}} & \lighthighghlighter &  \lighthighghlighter & \lighthighghlighter & \lighthighghlighter & \lighthighghlighter 
        \\

          \multirow{6}{*}{Mistral}
          & \multirow{3}{*}{LLM} & \multirow{3}{*}{Human} 
         &    Vanilla$^{\dagger}$ & 0.9647  & 0.9284
         \\ 
         & &
         & \highlighter Informed$^{\dagger}$ 
          &  \highlighter  0.9656   &  \highlighter 0.9257
         \\
         & & 
         &    CF-informed$^{\dagger}$ 
         & 0.9737
 &  0.9401
    \\ \cmidrule(r){2-6}
          & \multirow{3}{*}{Human} & \multirow{3}{*}{LLM} 
         &    Vanilla$^{\dagger}$ &  0.9678 & 0.9355
         \\ 
         & &
         & \highlighter Informed$^{\dagger}$ 
          &  \highlighter  0.9707   &  \highlighter 0.9400 
         \\
         & & 
         &    CF-informed$^{\dagger}$ 
         & 0.9638
 &  0.9434

\\ \midrule
          \multirow{6}{*}{Llama3}
         & \multirow{3}{*}{LLM} & \multirow{3}{*}{Human} 
         &    Vanilla$^{\dagger}$ & 0.9060  & 0.8145
         \\ 
         & &
         &  \highlighter Informed$^{\dagger}$ 
         &  \highlighter 0.8960   & \highlighter   0.8260
         
        \\ 
        & & 
        & 
           CF-informed$^{\dagger}$
        & 0.9235  & 0.8282
         
         \\ \cmidrule(r){2-6}
         & \multirow{3}{*}{Human} & \multirow{3}{*}{LLM} 
         &    Vanilla$^{\dagger}$ &  0.9088  & 0.7985       
         \\ 
         & &
         & \highlighter Informed$^{\dagger}$ 
          &  \highlighter 0.9163 &  \highlighter 0.8160 
         \\
         & & 
         &    CF-informed$^{\dagger}$ 
         &  0.8908
 &  0.8238


         \\ \midrule
          \multirow{6}{*}{GPT-4} 

         & \multirow{3}{*}{LLM} & \multirow{3}{*}{Human} 
         &    Vanilla$^{\dagger}$ &  0.9807 &  0.9042
         \\ 
         &  &
         & \highlighter Informed$^{\dagger}$ 
         & \highlighter 0.9796 &  \highlighter 0.9130
         \\ 
         &  & 
         &    CF-informed$^{\dagger}$
         &   0.9834  &  0.9094
         \\ \cmidrule(r){2-6}
         & \multirow{3}{*}{Human} & \multirow{3}{*}{LLM} 
         &    Vanilla$^{\dagger}$ & 0.9819  &  0.9238
         \\ 
         &  &
         & \highlighter Informed$^{\dagger}$ 
         & \highlighter 0.9778 &  \highlighter 0.9205   
         \\
         &  & 
         &    CF-informed$^{\dagger}$
         &   0.9776  &  0.9346

         \\ \bottomrule
    \end{tabular}
    }
    \caption{The attribution confidence (AC) of LLMs in
    relevant and non-relevant documents for NQ dataset. $\dagger$ indicates a statistically significant difference between the AC values of relevant and non-relevant documents.
    }
    \label{tab:confidence_table_NQ}
\end{table}
\header{Frequency of Attribution} 
In Table \ref{tab:groundness_table}, across the majority of the settings, \texttt{GPT-4} outperforms \texttt{Mistral} and \texttt{Llama3} in terms of precision of results. In contrast, in terms of recall, it is \texttt{Mistral} and \texttt{Llama3} which outperform \texttt{GPT-4}. To better explore this difference, we examine the average number of relevant citations and total citations for the three models. Figure \ref{fig:number-of-cited-docs} shows the average number of total citations\footnote{Tables \ref{tab:avg_table_NQ} and \ref{tab:avg_table_msmarco} in the Appendix show both the average number of relevant citations and the total citations.} for each model. In comparison to \texttt{Mistral} and \texttt{Llama3}, \texttt{GPT-4} tends to cite fewer documents as supporting documents for its generated answer.
This is in line with the previous results, where \texttt{Mistral} and \texttt{Llama3} score higher on recall. 

\header{Answer Correctness} Table \ref{tab:groundness_table} and~\ref{tab:brief_mixed_NQ} show that when the relevant documents are labeled with human-authorship and non-relevant ones are labeled with LLM-authorship, the answer correctness is higher than in the reverse case, across the majority of settings. Although this improvement is not significant and consistent across all settings, similar to attribution quality, it could indicate a bias towards human authorship. Nevertheless, we note that the automatic evaluation of answer correctness without human evaluation is not an ideal method \cite{chiang2023can,bojic2023hierarchical,zhang2019bertscore}.We leave this aspect for future work as the focus of this paper is on the performance of LLMs in how frequently they tend to cite and attribute their output on documents with either human or LLM authorship.
\begin{figure*}[t]
    \centering
    \includegraphics[width=\linewidth]{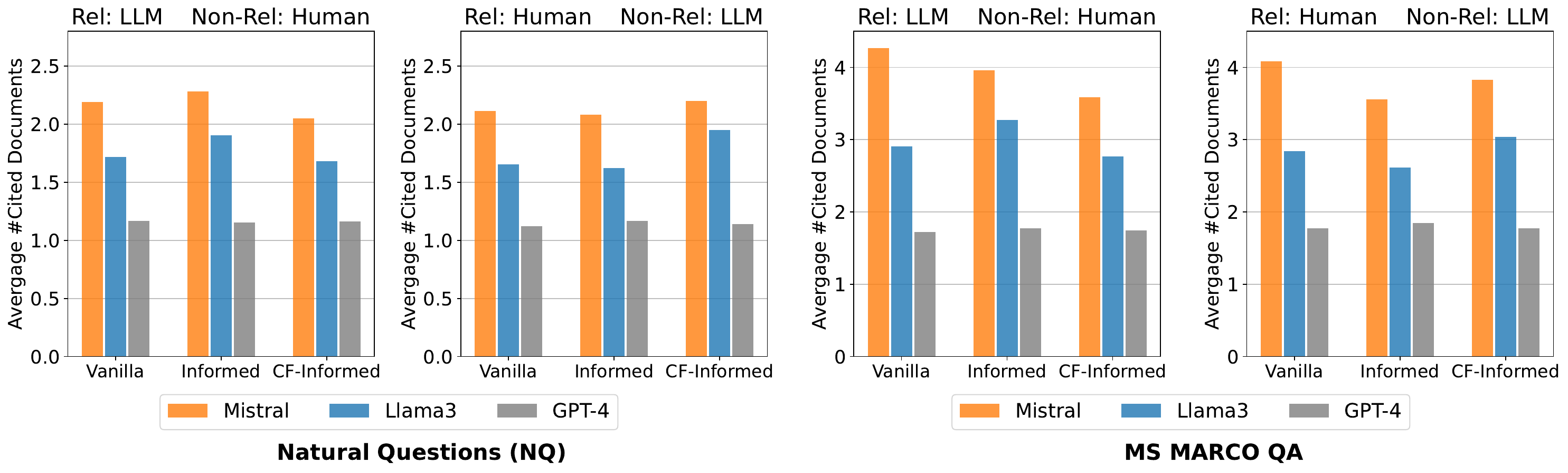}
    \caption{The average total number of cited documents by \texttt{Mistral}, \texttt{Llama3}, and \texttt{GPT-4} across various RAG settings on NQ and MS~MARCO benchmarks.}
    \label{fig:number-of-cited-docs}
\end{figure*}
\section{Conclusion and Future Work}
In this paper, we have defined and studied attribution sensitivity and bias with respect to authorship information of source documents in RAG with LLMs. 
We have proposed a systematic evaluation framework based on counterfactual evaluation.
Our results indicate that by adding authorship information to source documents, the attribution quality of LLMs may significantly change by 3\% to 18\%. 
Moreover, our results on three LLMs indicate that they have an attribution bias towards explicit human \emph{authorship}, in contrast to previous studies that show that LLM-generated \emph{content} may consistently be preferred over human-authored \emph{content} by LLMs. 

As to broader implications of our work, while understanding the roots and causes of the observed sensitivity and bias requires access to the implementation, training, and fine-tuning of these models (which is beyond the scope of this paper), our findings highlight a critical aspect of how LLMs operate. 
Our results show the brittleness of LLMs for attributing their answers. 
Such brittleness can be used for both constructive and harmful purposes, e.g., one can bias the output of an LLM towards their own content by incorporating authorship information in their documents. 

While we only focused on human versus LLM authorship as metadata in this work, in future work our systematic evaluation method can be used to investigate the sensitivity and bias towards other metadata of source documents (e.g., gender and race of authors).
Furthermore, our evaluation methodology can be incorporated in trustworthiness benchmarks used for the evaluation of LLMs such as DecodingTrust \cite{wang2024decodingtrust}.
Finally, our proposed methodology for the evaluation of sensitivity and bias is adaptable to other metrics for measuring the quality of attribution, i.e., metrics other than precision and recall can be used as $M$ in Eq. \ref{equation:cas}, \ref{equation:cab}, and~\ref{equation:cab_mixed}.

 
\section*{Limitations}
In this work we do not propose or explore solutions for mitigating the observed bias as our focus is on uncovering the brittleness of LLMs when being used for retrieval-augmented generation.
Besides, we have evaluated three LLMs in our experimental setup, two of which are open-source and the other closed-source. Consequently, investigating the same attribution sensitivity and bias on other LLMs is of interest for future studies. 
Furthermore, in our experiments, we used queries that have only one relevant document containing the ground-truth answer in their top-$k$ retrieved list of documents. 
We do this to ensure the traceability of the correct attribution. 
However, we acknowledge the limitation of this evaluation method in capturing the fine-grained attribution support of input documents.
Finally, it is important to mention that our current research is limited to datasets and prompts in English. Therefore, we point out the need to expand our evaluation and analysis to include datasets in other languages.

\bibliography{references}

\appendix

\section*{Appendix}
\section{Synthetic Document Generation}
\header{Prompt}
Figure \ref{tab:qagnostic_pp} shows the prompt used for re-writing passages for the two benchmarks of NQ and MS MARCO.
\vspace{4mm}
\begin{table}[ht]
\centering
\begin{tcolorbox}
[boxrule=0.3mm,
colback=gray!5,
                  colframe=gray, 
                  rounded corners, 
                  arc=0mm, 
                  ]
\fontsize{12}{13}\selectfont
\ttfamily{
\footnotesize
\textbf{Instruction}: Please write a high-quality paraphrase for the given passage.
\\Keep the length approximately the same. Do not add any new information. 
\\\\\textbf{Passage}: \{\textcolor{orange}{\textbf{input passage}}\}
}
\end{tcolorbox}

\captionof{figure}{Prompt used for generating synthetic documents.}
\label{tab:qagnostic_pp}
\end{table}
\header{Data Quality}
\label{appendix:data-quality}
In order to ensure the quality of synthetic passages, we conduct the following annotation steps using two expert annotators: (i) for each of the queries in our two benchmarks, we provide the annotators the quadruple of (query $q$, original relevant passage $p_r$, synthetic relevant $\hat{p}_r$ passage, answer $a$). We then ask the annotators to determine whether the synthetic passage $\hat{p}_r$ is still relevant to the query and includes the answer $a$ to the query $q$. (ii) In order to ensure that non-relevant passages are still non-relevant after being rewritten by an LLM, for each query, we provide the annotators the quadruple of (query $q$, original non-relevant passage $p_n$, synthetic non-relevant $\hat{p}_n$ passage, answer $a$). We then ask the annotators to determine whether the synthetic non-relevant passage $\hat{p}_n$ is still non-relevant to the query and does not include the answer $a$ to the query $q$. Due to the large number of non-relevant passages for each query, we randomly select 10\% of queries, i.e., 50 queries out of 500 queries. Our annotation results indicate a perfect performance in keeping the relevance and non-relevance status of synthetic documents with respect to their corresponding queries.
\section{Authorship Informed Answer/Attribution Generation Prompt}
Figure \ref{fig:informed_prompt} shows the prompt used for authorship-informed answer/attribution generation with all three LLMs, \texttt{Mistral}, \texttt{Llama3}, and \texttt{GPT-4}.

\begin{table*}[]
    \centering
    \setlength{\tabcolsep}{4.5pt}
\begin{tcolorbox}
[boxrule=0.2mm, colback=gray!2, 
                  colframe=darkgray, 
                  rounded corners, 
                  arc=0mm, 
                  ]
\fontsize{8.5}{10.5}\selectfont
\ttfamily{\textbf{Instruction}: Write a concise answer for the given question (query) based on the provided search result documents, and cite them properly using [0][1][2] etc. 
\\ \\Please take these strict considerations into account, during answer generation:
\\1. Documents are retrieved by a search engine. As such, not all the documents are relevant to the query. Only use and cite the relevant documents that contain the answer.
\\2. Do not analyze irrelevant documents.
\\3. Some documents are authored by humans, while others are generated by a Large Language Model (LLM). We denote this distinction for each document.
\\ \\ \textbf{Search Results}:
\\ \\ Document [0](\{\textcolor{orange}{\textbf{text of Document [0]}}\}) (written by \textcolor{orange}{\textbf{[Author]}})
\\  Document [1](\{\textcolor{orange}{\textbf{text of Document [1]}}\}) (written by \textcolor{orange}{\textbf{[Author]}})
\\ ...
\\  Document [9](\{\textcolor{orange}{\textbf{text of Document [9]}}\}) (written by \textcolor{orange}{\textbf{[Author]}})
\\ \\ \textbf{Question}: \{\textcolor{orange}{\textbf{query}}\}.
}
\end{tcolorbox}

    \captionof{figure}{Prompt used for authorship-informed answer/citation generation with LLM. \texttt{[Author]} is filled with one instance from either \{\texttt{Human}, \texttt{Person}, \texttt{Individual}\} or \{\texttt{AI}, \texttt{LLM}, \texttt{Machine}\}, depending on the source of the document and the RAG setting.
    }
    \label{fig:informed_prompt}
\end{table*}

\section{Extended Set of Authorship Labels}
\label{appendix:more-authorship-labels}
\amin{So far, we have used \texttt{[LLM]} and \texttt{[Human]} as the authorship labels for the source documents. In this section, we discuss and provide results using an extended set of authorship labels. Specifically, we use \texttt{[AI]} as the label for denoting the synthetic (LLM) authorship. For human authorship, on the other hand, we analyze the use of real-world names to indicate the authors of documents. This reflects a more realistic setting of authorship indication on documents. To create this set of names, we prompt \texttt{GPT-4} to randomly generate a pool of 100 (first name, last name) pairs. Figure \ref{prompt:namegeneration} shows the prompt we use for this task. We then randomly sample one instance of (first name, last name) from this pool when labeling human authorship for each document in the list of top-$k$ source documents of a query (instead of using \texttt{[Human]} as the authorship label).}
\begin{table}[ht]
\centering
\begin{tcolorbox}
[boxrule=0.3mm,
colback=gray!5,
                  colframe=gray, 
                  rounded corners, 
                  arc=0mm, 
                  ]
\fontsize{10}{12}\selectfont
\ttfamily{
\footnotesize
\textbf{Instruction}: Please generate a random list of 100 (first name, last name) pairs consisting of male and female names.
}
\end{tcolorbox}

\captionof{figure}{Prompt used for generating a pool of 100 pairs of (first name, last name).}
\label{prompt:namegeneration}
\end{table}

\amin{Table \ref{tab:haiwn_onlysensitivity} shows the attribution sensitivity results using the extended set of authorship labels. As we can see, all three LLMs \texttt{Mistral}, \texttt{Llama}, and \texttt{GPT-4} are sensitive to adding the authorship information similar to the attribution sensitivity results with \texttt{[Human]} and \texttt{[LLM]} authorship labels (Table \ref{tab:only_sensitivity}). In addition, we see that \texttt{GPT-4} shows a lower level of sensitivity than \texttt{Mistral} and \texttt{Llama}. Moreover, Table \ref{tab:haiwn_onlybias} shows the attribution bias results using the extended set of authorship labels. Similar to the attribution bias results with \texttt{[Human]} and \texttt{[LLM]} authorship labels (Table \ref{tab:only_bias}), all three LLMs \texttt{Mistral}, \texttt{Llama}, and \texttt{GPT-4} show an attribution bias towards human authorship, i.e., they are biased towards documents that are labeled with human author names. This indicates the robustness of our analysis against changes in labels.}

\begin{table}[ht]
    \centering
     \setlength{\tabcolsep}{2pt}
    \renewcommand{\arraystretch}{1}
    \scalebox{0.82}{
        \begin{tabular}{l ll cc}
        \toprule

        \multirow{2}{*}{\makecell{Answer \\ generator}} & \multirow{2}{*}{\makecell{Relevant \\ documents}} & \multirow{2}{*}{\makecell{Non-relevant \\ documents}}
         & \multicolumn{2}{c}{CAS} 
        \\ \cmidrule(r){4-5}
        &  & 
        & $\Delta$Precision & $\Delta$Recall

       \\ \midrule
       \multicolumn{1}{l}{\lighthighghlighter \footnotesize \texttt{NQ}} & \lighthighghlighter &  \lighthighghlighter & \lighthighghlighter & \lighthighghlighter
        \\
       \multirow{2}{*}{Mistral}  & \multirow{1}{*}{Human} & \multirow{1}{*}{Human} 

         & 27.5  & 26.8
         
         \\ 
         & \multirow{1}{*}{LLM} & \multirow{1}{*}{LLM} 

         & 13.3  & 14.4
         
         \\ \midrule 
         \multirow{2}{*}{Llama3} & \multirow{1}{*}{Human} & \multirow{1}{*}{Human} 

         &  15.0 &  12.4 
         \\ 
         & \multirow{1}{*}{LLM} & \multirow{1
         }{*}{LLM} 

         &  15.6 & 14.4
         \\ \cmidrule(r){1-5}

         \multirow{2}{*}{GPT-4} &  \multirow{1}{*}{Human} & \multirow{1}{*}{Human} 

         &  \phantom{0}7.4 &  \phantom{0}7.0

         \\ 
         & \multirow{1}{*}{LLM} & \multirow{1}{*}{LLM} 

         &  \phantom{0}7.5  &  \phantom{0}6.8

        \\ \midrule
        \multicolumn{1}{l}{\lighthighghlighter \footnotesize \texttt{MS MARCO}} & \lighthighghlighter &  \lighthighghlighter & \lighthighghlighter & \lighthighghlighter
        \\
       \multirow{2}{*}{Mistral}  & \multirow{1}{*}{Human} & \multirow{1}{*}{Human} 

         & 11.0 & 17.2 
         
         \\ 
         & \multirow{1}{*}{LLM} & \multirow{1}{*}{LLM} 

         & \phantom{0}9.4  & 14.0
         
         \\ \midrule 
        \multirow{2}{*}{Llama3}  & \multirow{1}{*}{Human} & \multirow{1}{*}{Human} 

         &  13.9  &  18.6
         
         \\ 
         & \multirow{1}{*}{LLM} & \multirow{1}{*}{LLM} 

         & 13.3 & 17.4
         
         \\ \midrule
          \multirow{2}{*}{GPT-4} &  \multirow{1}{*}{Human} & \multirow{1}{*}{Human} 

         
         &  10.8  &  13.2
         

         \\ 
         & \multirow{1}{*}{LLM} & \multirow{1}{*}{LLM} 

         &  \phantom{0}9.2  &  10.8

         \\ \bottomrule
    \end{tabular}
    }
    \caption{Attribution sensitivity (CAS) results for the RAG setting with extended set of authorship labels. Values range from 0 (minimum sensitivity) to 100 (maximum sensitivity). ${\dagger}$ indicates statistically significant values.}
    \label{tab:haiwn_onlysensitivity}
\end{table}
\begin{table}[ht]
    \centering
     \setlength{\tabcolsep}{2pt}
    \renewcommand{\arraystretch}{1}
    \scalebox{0.8}{
        \begin{tabular}{l ll cc}
        \toprule

        \multirow{2}{*}{\makecell{Answer \\ generator}} & \multirow{2}{*}{\makecell{Relevant \\ documents}} & \multirow{2}{*}{\makecell{Non-relevant \\ documents}}
         & \multicolumn{2}{c}{CAB} 
        \\ \cmidrule(r){4-5}
        &  & 
        & $\Delta$Precision & $\Delta$Recall

       \\ \midrule
       \multicolumn{1}{l}{\lighthighghlighter \footnotesize \texttt{NQ}} & \lighthighghlighter &  \lighthighghlighter & \lighthighghlighter & \lighthighghlighter
        \\
       \multirow{2}{*}{Mistral}  & \multirow{1}{*}{Human} & \multirow{1}{*}{Human} 

         &  +13.1 & \phantom{0}+3.6
         
         \\ 
         & \multirow{1}{*}{LLM} & \multirow{1}{*}{LLM} 

         &   +4.4   & \phantom{0}+2.4
         
         \\ \midrule 
         \multirow{2}{*}{Llama3} & \multirow{1}{*}{Human} & \multirow{1}{*}{Human} 

         & \phantom{0}+6.9  & \phantom{0}+1.6  
         \\ 
         & \multirow{1}{*}{LLM} & \multirow{1
         }{*}{LLM} 

         &  \phantom{0}+9.8 & \phantom{0}+8.4
         \\ \cmidrule(r){1-5}

         \multirow{2}{*}{GPT-4} &  \multirow{1}{*}{Human} & \multirow{1}{*}{Human} 

         & \phantom{0}+2.8  &  \phantom{0}+3.0

         \\ 
         & \multirow{1}{*}{LLM} & \multirow{1}{*}{LLM} 

         &  \phantom{0}+3.9  & \phantom{0}+2.4 

        \\ \midrule
        \multicolumn{1}{l}{\lighthighghlighter \footnotesize \texttt{MS MARCO}} & \lighthighghlighter &  \lighthighghlighter & \lighthighghlighter & \lighthighghlighter
        \\
       \multirow{2}{*}{Mistral}  & \multirow{1}{*}{Human} & \multirow{1}{*}{Human} 

         & \phantom{0}+6.6 &  \phantom{0}+6.0
         
         \\ 
         & \multirow{1}{*}{LLM} & \multirow{1}{*}{LLM} 

         & \phantom{0}+4.3  & \phantom{0}+3.6
         
         \\ \midrule 
        \multirow{2}{*}{Llama3}  & \multirow{1}{*}{Human} & \multirow{1}{*}{Human} 

         &  \phantom{0}+9.8  &  +12.2
         
         \\ 
         & \multirow{1}{*}{LLM} & \multirow{1}{*}{LLM} 

         &  \phantom{0}+8.0  & \phantom{0}+8.2
         
         \\ \midrule
          \multirow{2}{*}{GPT-4} &  \multirow{1}{*}{Human} & \multirow{1}{*}{Human} 

         
         &  \phantom{0}+5.1  &  \phantom{0}+4.0
         

         \\ 
         & \multirow{1}{*}{LLM} & \multirow{1}{*}{LLM} 

         &  \phantom{0}+6.9  &  \phantom{0}+6.8

         \\ \bottomrule
    \end{tabular}
    }
    \caption{Attribution Bias (CAB) results for the RAG setting with extended set of authorship labels. Positive values indicate a bias towards human. $\dagger$ indicates statistically significant bias values.}
    \label{tab:haiwn_onlybias}
\end{table}
\section{Effect of the Number of Source Documents}
\label{app:topk-effect}
\begin{figure*}[ht]
    \centering
    \includegraphics[width=0.95\linewidth]{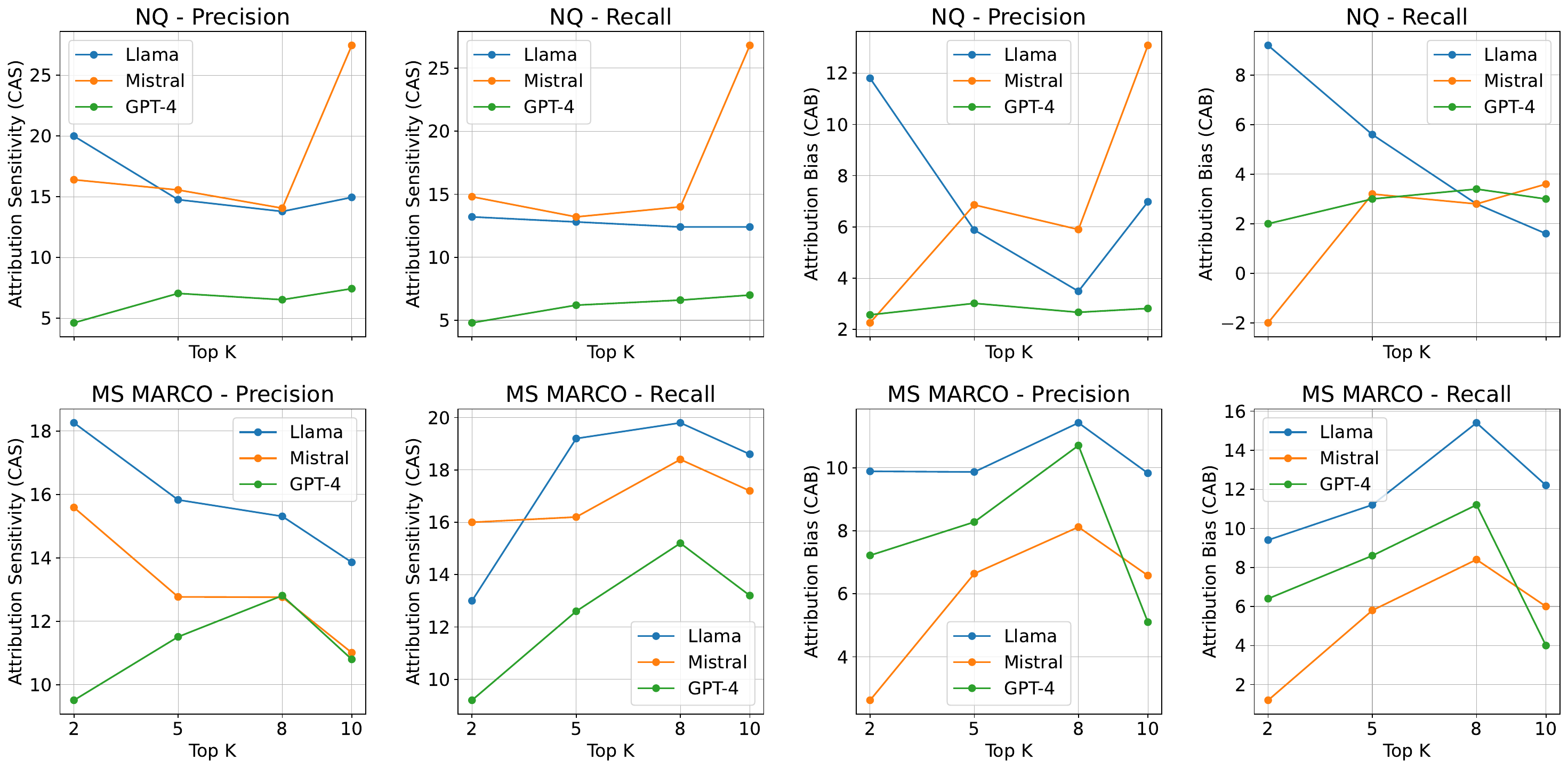}
    \caption{Attribution Sensitivity and Bias in  \texttt{Mistral}, \texttt{Llama3}, and \texttt{GPT-4}, across varying number of retrieved documents (top-$k$ values) on NQ (top) and MS MARCO benchmarks (bottom).}
    \label{fig:topk-effect}
\end{figure*}

\amin{To study the effect of the number of source documents, i.e., the length of the retrieved ranked list of documents given to the answer generator LLM, we evaluate the attribution sensitivity and bias using varying number of source documents. To this end, we use 4 ranking cut-offs for the ranked list of source documents ($k$): 2, 5, 8, 10. To ensure the existence of the relevant document as the input, we randomly put the relevant document in the top-$k$ ($k \in \{2,5,8,10\}$). For this set of experiments we use human-generated versions of both relevant and non-relevant documents. Furthermore, we use the extended set of labels (i.e., authors with first names and last names). 
Figure \ref{fig:topk-effect} shows the results of attribution sensitivity (CAS) and attribution bias (CAB) for the three LLMs on the NQ and MS MARCO benchmarks. All three LLMs show both attribution sensitivity and bias across varying number of source documents ($k$). Moreover, we can see that no conclusion can be inferred for the effect of $k$ on the \emph{degree} of sensitivity and bias in these LLMs.}

\section{Effect of the Retriever}
\label{app:retrievers}
\amin{In our experiments, we have used two different retrievers for NQ and MS MARCO benchmarks: the list of source documents for NQ are retrieved using BM25, and for MS MARCO we used the ranked list of documents in the benchmark which are retrieved using the Bing search engine (see Section \ref{sec:experimental-settings}).}

\amin{In order to better disentangle the effect of retrievers on the attribution sensitivity and bias, we use two more commonly-used retrievers for our experiments:}
\begin{itemize}[leftmargin=*,nosep]
    \item \texttt{uniCOIL} \cite{lin2021unicoil}: a retrieval model built upon COIL~\cite{gao2021coil}, which works based on sparse learned representation of documents.
    
    \item \texttt{TCT-ColBERT} \cite{lin2020tctcolbertv1}: a dense retrieval model trained with knowledge distillation using ColBERT~\cite{khattab2020colbert} as the teacher model.
\end{itemize}
\amin{For this set of experiments we use the extended set of labels. Besides, we use original (human-generated) documents. Table \ref{tab:haiwn_retrievers} shows the results of attribution sensitivity and bias on NQ benchmark using \texttt{uniCOIL} and \texttt{TCT-ColBERT}. As the results on \texttt{uniCOIL} and \texttt{TCT-ColBERT} show, the three LLMs \{\texttt{Mistral}, \texttt{Llama}, \texttt{GPT-4}\} have attribution sensitivity and bias with respect to the authorship information regardless of the retriever that is being used to retrieve their top-$k$ source documents. Moreover, we see that the sensitivity and bias values across all models are lower for the answer generation upon the source documents from \texttt{uniCOIL} than when \texttt{TCT-ColBERT} is being used as the retriever. This finding is specifically important as it shows that the quality of retrieved source documents can affect the quality of attribution by LLMs.}
\begin{table}[t]
    \centering
    \renewcommand{\arraystretch}{1}
    \scalebox{0.83}{
        \begin{tabular}{l l cc}
        \toprule

        \multirow{2}{*}{\makecell{Answer \\ generator}} & \multirow{2}{*}{Retriever}  & \multirow{2}{*}{$\Delta$Precision} & \multirow{2}{*}{$\Delta$Recall}
        \\ 
        & 
        &  & 

       \\ \midrule
       \multicolumn{1}{l}{\lighthighghlighter \footnotesize \texttt{\textbf{CAS}}} & \lighthighghlighter &  \lighthighghlighter & \lighthighghlighter
        \\
       \multirow{2}{*}{Mistral}  & \multirow{1}{*}{uniCOIL} 

         & 16.8 & 15.0
         
         \\ 
         & \multirow{1}{*}{TCT-ColBERT}

         &  18.2   & 17.0
         
         \\ \midrule 
         \multirow{2}{*}{Llama3} & \multirow{1}{*}{uniCOIL} 

         & 14.5  & 13.0 
         \\ 
         & \multirow{1}{*}{TCT-ColBERT} 

         & 18.0  & 13.6
         \\ \cmidrule(r){1-4}

         \multirow{2}{*}{GPT-4} &  \multirow{1}{*}{uniCOIL}

         & \phantom{0}6.6 & \phantom{0}6.6 

         \\ 
         & \multirow{1}{*}{TCT-ColBERT}

         & \phantom{0}8.7   & \phantom{0}8.2

        \\ \midrule
        \multicolumn{1}{l}{\lighthighghlighter \footnotesize \texttt{\textbf{CAB}}} &  \lighthighghlighter & \lighthighghlighter & \lighthighghlighter
        \\
       \multirow{2}{*}{Mistral}  & \multirow{1}{*}{uniCOIL} 
         & \phantom{0}+6.6 &  \phantom{0}+3.4
         
         \\ 
         & \multirow{1}{*}{TCT-ColBERT}

         & \phantom{0}+7.9 & \phantom{0}+5.8
         
         \\ \midrule 
        \multirow{2}{*}{Llama3}  & \multirow{1}{*}{uniCOIL} 
         &  \phantom{0}+8.2  & \phantom{0}+4.6 
         
         \\ 
         & \multirow{1}{*}{TCT-ColBERT} 

         & +12.7 & \phantom{0}+8.8
         
         \\ \midrule
          \multirow{2}{*}{GPT-4} &  \multirow{1}{*}{uniCOIL} 

         
         & \phantom{0}+3.9 &  \phantom{0}+3.8
         

         \\ 
         & \multirow{1}{*}{TCT-ColBERT} 
         & \phantom{0}+5.2  & \phantom{0}+4.6 

         \\ \bottomrule
    \end{tabular}
    }
    \caption{Attribution sensitivity (CAS) and Bias (CAB) results across different retrievers. Positive values of CAB indicate a bias towards human authorship.}
    \label{tab:haiwn_retrievers}
\end{table}
\begin{table}[ht]
    \centering
        \renewcommand{\arraystretch}{1.1}

    \setlength{\tabcolsep}{1.2pt}
    \scalebox{0.7
    }{
    \begin{tabular}{l ll l cc }
        \toprule

        \multirow{2}{*}{\makecell{Answer \\ generator}} & \multirow{2}{*}{\makecell{Relevant \\ documents}} & \multirow{2}{*}{\makecell{Non-relevant \\ documents}}
        & \multirow{2}{*}{\makecell{RAG \\ mode}} &  \multicolumn{2}{c}{Confidence} 
        \\ \cmidrule(r){5-6} 
        &  & 
        &  & Relevant &  Non-relevant 
        \\ \midrule
        \multicolumn{1}{l}{\lighthighghlighter \footnotesize \texttt{MS MARCO}} & \lighthighghlighter &  \lighthighghlighter & \lighthighghlighter & \lighthighghlighter & \lighthighghlighter

                \\ 
   \multirow{6}{*}{Mistral} 

        \multirow{6}{*}{}  & \multirow{3}{*}{LLM} & \multirow{3}{*}{Human} 
         &    Vanilla &  0.9620 & 0.9527
         \\ 
         & &
         &  \highlighter  Informed 
         & \highlighter 0.9511 & \highlighter 0.9470
        \\ 
        & & 
        & 
          CF-informed$^{\dagger}$
        &  0.9746 &  0.9456
         
         \\ \cmidrule(r){2-6}
         & \multirow{3}{*}{Human} & \multirow{3}{*}{LLM} 
         &    Vanilla$^{\dagger}$ & 0.9616 & 0.9446
         
         \\ 
         & &
         & \highlighter Informed 
          &  \highlighter 0.9650 &  \highlighter 0.9521
         \\ 
         & & 
         &    CF-Informed 
         & 0.9484   &   0.9516      

                \\ \midrule
        \multirow{6}{*}{Llama3}

        \multirow{6}{*}{}  & \multirow{3}{*}{LLM} & \multirow{3}{*}{Human} 
         &    Vanilla$^{\dagger}$ &  0.9267  &  0.8878
         \\ 
         & &
         &  \highlighter  Informed$^{\dagger}$ 
         & \highlighter 0.9104 & \highlighter 0.8918
        \\ 
        & & 
        & 
          CF-informed$^{\dagger}$
        & 0.9332  &  0.8622
         
         \\ \cmidrule(r){2-6}
         & \multirow{3}{*}{Human} & \multirow{3}{*}{LLM} 
         &    Vanilla & 0.8888 &  0.8941
         
         \\ 
         & &
         & \highlighter Informed$^{\dagger}$ 
          &  \highlighter 0.9441 &  \highlighter  0.8736
         \\ 
         & & 
         &    CF-Informed$^{\dagger}$ 
         &  0.906  &   0.889
         
         \\ \midrule
          \multirow{6}{*}{GPT-4} 
   
         & \multirow{3}{*}{LLM} & \multirow{3}{*}{Human} 
         &    Vanilla$^{\dagger}$ & 0.9749  &  0.9038
         \\ 
         &  &
         & \highlighter Informed$^{\dagger}$ 
         & \highlighter 0.9714  &  \highlighter 0.9165
         \\ 
         &  & 
         &    CF-informed$^{\dagger}$
         &  0.9757 & 0.9173

         \\ \cmidrule(r){2-6}
         & \multirow{3}{*}{Human} & \multirow{3}{*}{LLM} 
         &    Vanilla &  0.9506 &  0.9395
         \\ 
         &  &
         & \highlighter Informed$^{\dagger}$ 
         & \highlighter 0.9657  &  \highlighter  0.9171 
         \\ 
         &  & 
         &    CF-informed$^{\dagger}$
         &   0.9556 &  0.936\phantom{0}

         \\ \bottomrule
    \end{tabular}
    
    }
    \caption{The attribution confidence (AC) of LLMs in attributing answers to relevant and non-relevant documents for the MS MARCO QA benchmark. $\dagger$ stands for statistically significant difference between the AC values of relevant and non-relevant documents.}
    \label{tab:confidence_table_NLG}
\end{table}
\section{Attribution Quality Results}
Table \ref{tab:attributionquality-msmarco} shows the results of attribution by \texttt{Mistral}, \texttt{Llama3}, and \texttt{GPT-4}, under different settings for the MS MARCO benchmark.
\begin{table*}[ht]
    \centering
    \renewcommand{\arraystretch}{1}
    \scalebox{0.8
    }{
    \begin{tabular}{l ll l ccc }
        \toprule

        \multirow{2}{*}{\makecell{Answer \\ generator}} & \multirow{2}{*}{\makecell{Relevant \\ documents}} & \multirow{2}{*}{\makecell{Non-relevant \\ documents}}
        & \multirow{2}{*}{\makecell{RAG \\ mode}} &  \multicolumn{2}{c}{Attribution quality} & Correctness 
        \\ \cmidrule(r){5-6} \cmidrule(r){7-7}  
        &  & 
        &  & Precision &  Recall  & EM
        \\ \midrule
        \multicolumn{1}{l}{\lighthighghlighter \footnotesize \texttt{MS MARCO}} & \lighthighghlighter &  \lighthighghlighter & \lighthighghlighter & \lighthighghlighter & \lighthighghlighter & \lighthighghlighter 
        \\
     \multirow{6}{*}{Mistral} &  \multirow{3}{*}{LLM} & \multirow{3}{*}{Human} 
         &    Vanilla  & 23.1 &  76.4 & \textbf{0.316}
         \\ 
         & &
         & \highlighter Informed 
       & \highlighter 22.2
  & \highlighter 65.8  & \highlighter 0.306 
         \\ 
         & &
         &    CF-informed
         &    \textbf{31.7}\rlap{$^\dagger$} &    \textbf{79.6}\rlap{$^\dagger$} &      0.312

         \\ \cmidrule(r){2-7}
         & \multirow{3}{*}{Human} & \multirow{3}{*}{LLM} 
         &  Vanilla  & 22.8 & 72.8  & 0.342 
         \\ 
         &  &
         & \highlighter Informed 
            & \highlighter \textbf{28.0}\rlap{$^\dagger$} & \highlighter \textbf{72.6}\rlap{$^\dagger$} & \highlighter \textbf{0.384}
         \\ 
         &  & 
         &    CF-informed
        & 20.1
 & 60.2
 & 0.334
        \\ \midrule
        \multirow{6}{*}{Llama3}  & \multirow{3}{*}{LLM} & \multirow{3}{*}{Human} 
         &    Vanilla &    29.3 &    66.0  
          &    0.334
         \\ 
         & &
         &  \highlighter  Informed 
         & \highlighter  22.8 & \highlighter  58.0  
        & \highlighter 0.330
        \\ 
        & & 
        & 
          CF-informed
        &    \textbf{38.4}\rlap{$^\dagger$} &    \textbf{76.2}\rlap{$^\dagger$} 
        &    \textbf{0.352}
         
         \\ \cmidrule(r){2-7}
         & \multirow{3}{*}{Human} & \multirow{3}{*}{LLM} 
         &    Vanilla &    30.5 &    64.8  
         
          &    0.416
         
         \\ 
         & &
         & \highlighter Informed 
          &  \highlighter \textbf{42.6}\rlap{$^\dagger$} &  \highlighter \textbf{78.0}\rlap{$^\dagger$} &   \highlighter \textbf{0.474}
         \\ 
         & & 
         &    CF-Informed 
         &   27.5
 &    61.6
 &      0.422

         \\ \midrule

          \multirow{6}{*}{GPT-4} &  \multirow{3}{*}{LLM} & \multirow{3}{*}{Human} 
         &    Vanilla &    38.1 &    55.6 
         &    0.312
         \\ 
         & &
         & \highlighter Informed 
         & \highlighter 35.4 & \highlighter \highlighter 52.0 
         & \highlighter 0.310
         \\ 
         & &
         &    CF-informed
         &    \textbf{41.5}\rlap{$^\dagger$} &    \textbf{61.0}\rlap{$^\dagger$} &        \textbf{0.324}

         \\ \cmidrule(r){2-7}
         & \multirow{3}{*}{Human} & \multirow{3}{*}{LLM} 
         &    Vanilla &    37.0 &    53.0 
         &    \textbf{0.380}
         \\ 
         &  &
         & \highlighter Informed 
         & \highlighter \textbf{38.5} &  \highlighter \textbf{59.2}\rlap{$^\dagger$} & \highlighter 0.378
         \\ 
         &  & 
         &    CF-informed
         &    33.1 &    48.4
         &    0.362 

         \\ \bottomrule
    \end{tabular}
    
    }
    \caption{Quality of attribution and answer correctness for MS MARCO. The columns ``Relevant Documents'' and ``Non-relevant Documents'' refer to the actual authorship of input documents. Informed refers to the authorship-informed RAG and CF-informed refers to counterfactual-authorship informed RAG (Section \ref{subsec:ragmodes}). $\dagger$ indicates statistically significant improvements over the two other RAG Modes in each combination of ``Relevant'' and ``Non-relevant'' documents.
    }    
    \label{tab:attributionquality-msmarco}
\end{table*}
\section{Confidence Results}
Table \ref{tab:confidence_table_NLG} shows the confidence results of \texttt{Mistral}, \texttt{Llama3}, and \texttt{GPT-4} on MS MARCO benchmark.

\section{Average Number of Cited Documents}
Tables \ref{tab:avg_table_NQ} and \ref{tab:avg_table_msmarco} show \emph{Relevant} and \emph{Total} number of cited documents for each model on both benchmarks.
\begin{table}[ht]
    \centering
    \setlength{\tabcolsep}{2.8pt}
    \scalebox{0.7}{
    \begin{tabular}{l ll l cc }
        \toprule

        \multirow{2}{*}{\makecell{Answer \\ generator}} & \multirow{2}{*}{\makecell{Relevant \\ documents}} & \multirow{2}{*}{\makecell{Non-relevant \\ documents}}
        & \multirow{2}{*}{\makecell{RAG \\ mode}} &  \multicolumn{2}{c}{\#Cited docs.} 
        \\ \cmidrule(r){5-6} 
        &  & 
        &  & Relevant &  Total 
        \\ \midrule
       \multicolumn{1}{l}{\lighthighghlighter \footnotesize \texttt{NQ}} & \lighthighghlighter &  \lighthighghlighter & \lighthighghlighter & \lighthighghlighter & \lighthighghlighter 
        \\
        \multirow{6}{*}{Mistral}  & \multirow{3}{*}{LLM} & \multirow{3}{*}{Human} 
         &    Vanilla &  0.766  &  2.190
         \\ 
         & &
         &  \highlighter  Informed 
         & \highlighter 0.682  & \highlighter  2.280  
        \\ 
        & & 
        & 
          CF-informed
        &   0.778  & 2.050 
         
         \\ \cmidrule(r){2-6}
         & \multirow{3}{*}{Human} & \multirow{3}{*}{LLM} 
         &    Vanilla &  0.784  & 2.114
         
         \\ 
         & &
         & \highlighter Informed 
          &  \highlighter 0.778  &  \highlighter 2.080 
         \\ 
         & & 
         &    CF-Informed 
         &   0.702 &  2.202
         \\ \midrule
         \multirow{6}{*}{Llama3} & \multirow{3}{*}{LLM} & \multirow{3}{*}{Human} 
         &    Vanilla &  0.692  & 1.718
         \\ 
         & &
         &  \highlighter Informed 
         &  \highlighter 0.696   & \highlighter 1.906    
        \\ 
        & & 
        & 
           CF-informed
        & 0.776  & 1.682
         
         \\ \cmidrule(r){2-6}
         & \multirow{3}{*}{Human} & \multirow{3}{*}{LLM} 
         &    Vanilla &  0.710  & 1.656  
         \\ 
         & &
         & \highlighter Informed 
          &  \highlighter 0.778 &  \highlighter 1.624
         \\
         & & 
         &    CF-informed 
         &   0.692
 &  1.952
         
         \\ \cmidrule(r){1-6}

         \multirow{6}{*}{GPT-4} &  \multirow{3}{*}{LLM} & \multirow{3}{*}{Human} 
         &    Vanilla & 0.688 &  1.166
         \\ 
         & &
         & \highlighter Informed 
         &  \highlighter 0.646  & \highlighter 1.152 
         \\ 
         & &
         &    CF-informed
         & 0.722   & 1.162
         \\ \cmidrule(r){2-6}
         & \multirow{3}{*}{Human} & \multirow{3}{*}{LLM} 
         &  Vanilla & 0.688  &  1.122 
         \\ 
         &  &
         &  \highlighter  Informed 
         &  \highlighter 0.722 &  \highlighter 1.168  
         \\ 
         &  & 
         &  CF-informed
                         &  0.650  &   1.138

         \\ \bottomrule
    \end{tabular}
    
    }
    \caption{The average number of cited relevant documents and in total (relevant plus non-relevant documents).}
    \label{tab:avg_table_NQ}
\end{table}
\begin{table}[ht]
    \centering
    \setlength{\tabcolsep}{2.8pt}
    \scalebox{0.7}{
    \begin{tabular}{l ll l cc }
        \toprule

        \multirow{2}{*}{\makecell{Answer \\ generator}} & \multirow{2}{*}{\makecell{Relevant \\ documents}} & \multirow{2}{*}{\makecell{Non-relevant \\ documents}}
        & \multirow{2}{*}{\makecell{RAG \\ mode}} &  \multicolumn{2}{c}{\#Cited docs.} 
        \\ \cmidrule(r){5-6} 
        &  & 
        &  & Relevant &  Total 
        \\ \midrule
        \multicolumn{1}{l}{\lighthighghlighter \footnotesize \texttt{MS MARCO}} & \lighthighghlighter &  \lighthighghlighter & \lighthighghlighter & \lighthighghlighter & \lighthighghlighter

           \\
        \multirow{6}{*}{Mistral}  & \multirow{3}{*}{LLM} & \multirow{3}{*}{Human} 
         &    Vanilla &   0.764 &  4.266
         \\ 
         & &
         &  \highlighter  Informed 
         & \highlighter  0.658 & \highlighter 3.960  
        \\ 
        & & 
        & 
          CF-informed
        &  0.796       &   3.586
         
         \\ \cmidrule(r){2-6}
         & \multirow{3}{*}{Human} & \multirow{3}{*}{LLM} 
         &    Vanilla &  0.728  & 4.084
         
         \\ 
         & &
         & \highlighter Informed 
          &  \highlighter 0.726   &  3.560 \highlighter 
         \\ 
         & & 
         &    CF-Informed 
         & 0.602   &  3.826
         \\ \midrule
        \multirow{6}{*}{Llama3}  & \multirow{3}{*}{LLM} & \multirow{3}{*}{Human} 
         &    Vanilla &  0.66\phantom{0}  &  2.91\phantom{0}
         \\ 
         & &
         &  \highlighter  Informed 
         & \highlighter 0.58\phantom{0}  & \highlighter 3.274  
        \\ 
        & & 
        & 
          CF-informed
        &  0.762   &  2.77\phantom{0} 
         
         \\ \cmidrule(r){2-6}
         & \multirow{3}{*}{Human} & \multirow{3}{*}{LLM} 
         &    Vanilla &  0.648  &  2.838
         
         \\ 
         & &
         & \highlighter Informed 
          &  \highlighter 0.78\phantom{0} &  \highlighter 2.614 
         \\ 
         & & 
         &    CF-Informed 
         &  0.616  &   3.038
         \\ \midrule

          \multirow{6}{*}{GPT-4} &  \multirow{3}{*}{LLM} & \multirow{3}{*}{Human} 
         &    Vanilla & 0.556   &  1.724 
         \\ 
         & &
         & \highlighter Informed 
         & \highlighter 0.52\phantom{0}  & \highlighter  1.774  
         \\ 
         & &
         &  CF-informed
         &  0.61\phantom{0}  &  1.744
         
         \\ \cmidrule(r){2-6}
         & \multirow{3}{*}{Human} & \multirow{3}{*}{LLM} 
         &    Vanilla &  0.53\phantom{0} &  1.772 
         \\ 
         &  &
         & \highlighter Informed 
         & \highlighter 0.592 &  \highlighter 1.848
         \\ 
         &  & 
         &    CF-informed
         &  0.484   &  1.776
         \\ \bottomrule
    \end{tabular}
    
    }
    \caption{The average number of cited relevant documents and in total (relevant plus non-relevant documents).}
    \label{tab:avg_table_msmarco}
\end{table}
\section{Mixed RAG Mode Results}
Tables \ref{tab:mixed_full1} and \ref{tab:mixed_full2} show the results for Mixed RAG mode as described in Section~\ref{sec:mixed-rag-mode}. 

\begin{table*}[ht]
    \centering
   \setlength{\tabcolsep}{2.7pt}
   \renewcommand{\arraystretch}{1}

      \begin{tabular}{l ll ll ccc}
        \toprule

        \multirow{2}{*}{\makecell{Answer \\ generator}} & \multirow{2}{*}{\makecell{Relevant \\ documents}} & \multirow{2}{*}{\makecell{Non-relevant \\ documents}}
        & \multicolumn{2}{c}{Mixed RAG mode} &  \multicolumn{2}{c}{Attribution quality} & Correctness 
        \\ \cmidrule(r){4-5} \cmidrule(r){6-7} \cmidrule(r){8-8} 
        &  & & Relevant
        & Non-relevant  & Precision &  Recall &  EM
        \\ \midrule

        \multicolumn{1}{l}{\lighthighghlighter \footnotesize \texttt{NQ}} & \lighthighghlighter &  \lighthighghlighter & \lighthighghlighter & \lighthighghlighter & \lighthighghlighter & \lighthighghlighter &
        \lighthighghlighter

          \\
        \multirow{10}{*}{Mistral}& \multirow{5}{*}{Human} & \multirow{5}{*}{Human} & 
        
        Vanilla & Vanilla  & 50.4 & \textbf{77.6}  & \textbf{0.784}   
        \\
        & & & 
         \highlighter Informed &  \highlighter Informed &  \highlighter 45.5 & \highlighter74.6 &\highlighter 0.772
        \\
        & & &
        CF-informed & Informed&  44.8 &  71.8 
        &   0.772  

        \\
        &&& 
            \highlighter Informed &  \highlighter CF-informed & \highlighter
\textbf{52.3}  &  \highlighter 77.2
 & \highlighter  0.780 
        \\
        &&&
        CF-informed & CF-informed  &  46.3 &  73.2 & 0.768

        \\ \cmidrule(r){2-8}
        & \multirow{5}{*}{LLM} & \multirow{5}{*}{LLM} & 
        
        Vanilla & Vanilla  & 47.0
 & \textbf{76.8}
  & 0.724
        \\
        & & & 
         \highlighter Informed &  \highlighter Informed & \highlighter 48.4
  & \highlighter 74.6
  & \highlighter 0.726
        \\
        & & &
        CF-informed & Informed &  \textbf{48.7} &  74.6 
        &   0.718
        \\
        &&& 
         \highlighter Informed &  \highlighter CF-informed & \highlighter 42.9
  &  \highlighter 69.4
 & \highlighter \textbf{0.742}

        \\
        &&&
        CF-informed & CF-informed  &  46.0
 &  72.6 & 0.740

        \\ \midrule
        \multirow{10}{*}{Llama3}& \multirow{5}{*}{Human} & \multirow{5}{*}{Human} & 
        
        Vanilla & Vanilla  & 50.4 & 72.0 &  0.798
        \\
        & & & 
         \highlighter Informed &  \highlighter Informed &  \highlighter 46.6  &  \highlighter 71.0 &  \highlighter 0.796
        \\
        & & &
        CF-informed & Informed &  45.7 & 69.6  & 0.784    

        \\
        &&& 
            \highlighter Informed &  \highlighter CF-informed &  \highlighter \textbf{57.4} &   \highlighter \textbf{77.6 } &  \highlighter \textbf{0.808}
        \\
        &&&
        CF-informed & CF-informed & 48.8  & 69.2  & 0.780

        \\ \cmidrule(r){2-8}
        & \multirow{5}{*}{LLM} & \multirow{5}{*}{LLM} & 
        
        Vanilla & Vanilla  & 53.1 & 71.4  & 0.742
        \\
        & & & 
         \highlighter Informed &  \highlighter Informed &  \highlighter 50.4 &  \highlighter 68.8  &  \highlighter 0.732
        \\
        & & &
        CF-informed & Informed &  \textbf{59.3} &  \textbf{77.8} 
        & \textbf{0.744}
        \\
        &&& 
         \highlighter Informed &  \highlighter CF-informed &  \highlighter 44.7
 &  \highlighter 68.4
 &  \highlighter 0.726

        \\
        &&&
        CF-informed & CF-informed  & 50.8 & 75.8  & 0.732
        
        \\ \midrule
        \multirow{10}{*}{GPT-4}& \multirow{5}{*}{Human} & \multirow{5}{*}{Human} & 
        
        Vanilla & Vanilla & 65.9 & 71.2 & 0.778    
        \\
        & & & 
         \highlighter Informed &  \highlighter Informed &  \highlighter 68.1 &  \highlighter 73.2  &  \highlighter 0.786 
        \\
        & & &
        CF-informed & Informed &  65.8 & 70.6 & \textbf{0.794}
        \\
        &&& 
         \highlighter Informed &  \highlighter CF-informed &  \highlighter \textbf{69.1} &   \highlighter \textbf{74.0} &  \highlighter 0.784
        \\
        &&&
        CF-informed & CF-informed & 66.9 & 72.6  & 0.790
        
        \\ \cmidrule(r){2-8}
        & \multirow{5}{*}{LLM} & \multirow{5}{*}{LLM} & 
        
        Vanilla & Vanilla & 65.9 & 70.4   & 0.718
        \\
        & & & 
         \highlighter Informed &  \highlighter Informed &   \highlighter 65.2 &  \highlighter 69.8 &   \highlighter 0.726
        \\
        & & &
        CF-informed & Informed &\textbf{ 66.1} & \textbf{71.2}   & \textbf{0.730}
        \\
        &&& 
         \highlighter Informed &  \highlighter CF-informed &  \highlighter 61.7 &  \highlighter 66.8 &  \highlighter 0.716
        \\
        &&&
        CF-informed & CF-informed & 63.8 & 68.8  & 0.724

         \\ \bottomrule
    \end{tabular}
    
    \caption{Quality of attribution and answer correctness with Mixed RAG modes for NQ benchmark. The columns ``Relevant Documents'' and ``Non-relevant Documents'' refer to the actual authorship of input documents.
    }
    \label{tab:mixed_full1}
\end{table*}
\begin{table*}[ht]
    \centering
   \setlength{\tabcolsep}{2.7pt}
   \renewcommand{\arraystretch}{1}
      \begin{tabular}{l ll ll ccc}
        \toprule

        \multirow{2}{*}{\makecell{Answer \\ generator}} & \multirow{2}{*}{\makecell{Relevant \\ documents}} & \multirow{2}{*}{\makecell{Non-relevant \\ documents}}
        & \multicolumn{2}{c}{Mixed RAG mode} &  \multicolumn{2}{c}{Attribution quality} & Correctness 
        \\ \cmidrule(r){4-5} \cmidrule(r){6-7} \cmidrule(r){8-8} 
        &  & & Relevant
        & Non-relevant  & Precision &  Recall &  EM
        \\ \midrule
        \multicolumn{1}{l}{\lighthighghlighter \footnotesize \texttt{MS MARCO QA}} & \lighthighghlighter &  \lighthighghlighter & \lighthighghlighter & \lighthighghlighter & \lighthighghlighter & \lighthighghlighter &
        \lighthighghlighter
                  \\
        \multirow{10}{*}{Mistral}& \multirow{5}{*}{Human} & \multirow{5}{*}{Human} & 
        
        Vanilla & Vanilla  & 22.7 & 75.6 & 0.370
        \\
        & & & 
         \highlighter Informed &  \highlighter Informed  & \highlighter 22.7 & \highlighter 71.6 & \highlighter 0.368
        \\
        & & &
        CF-informed & Informed &  19.8
  &  62.4
        &  0.370   

        \\
        &&& 
            \highlighter Informed &  \highlighter CF-informed  &
 \highlighter \textbf{28.4} &  \highlighter \textbf{77.2}
 & \highlighter \textbf{0.389}
        \\
        &&&
        CF-informed & CF-informed  &  24.4 & 71.6  & 0.380

        \\ \cmidrule(r){2-8}
        & \multirow{5}{*}{LLM} & \multirow{5}{*}{LLM} & 
        
        Vanilla & Vanilla & 24.0
 & 73.6
  & 0.298   
        \\
        & & & 
         \highlighter Informed &  \highlighter Informed & \highlighter 23.6
 & \highlighter 61.8 & \highlighter 0.298
        \\
        & & &
        CF-informed & Informed & \textbf{ 28.9}  & \textbf{ 75.6 }
        &   0.296
        \\
        &&& 
         \highlighter Informed &  \highlighter CF-informed  &  \highlighter 20.2
 & \highlighter 61.8 & \highlighter 0.278

        \\
        &&&
        CF-informed & CF-informed   &  23.3 & 70.8
 & 0.276
        
        \\ \midrule
        \multirow{10}{*}{Llama3}& \multirow{5}{*}{Human} & \multirow{5}{*}{Human} & 
        
        Vanilla & Vanilla & 30.4 & 70.0 & 0.436
        \\
        & & & 
        \highlighter Informed & \highlighter Informed & \highlighter 29.9 & \highlighter 74.4 & \highlighter 0.430
        \\
        & & &
        CF-informed & Informed &   24.9 & 70.0  & 0.432
        \\
        &&& 
        \highlighter Informed & \highlighter CF-informed &  \highlighter \textbf{37.5} &  \highlighter \textbf{80.4}  &  \highlighter \textbf{0.476}
        
        \\
        &&&
        CF-informed & CF-informed & 28.8 & 66.8 
        & 0.424
        \\ \cmidrule(r){2-8}
        & \multirow{5}{*}{LLM} & \multirow{5}{*}{LLM} & 
        
        Vanilla & Vanilla & 30.1 & 65.2
         & 0.326
        \\
        & & & 
         \highlighter Informed &  \highlighter Informed &  \highlighter 31.5 &  \highlighter 65.6 
         &  \highlighter 0.330
        \\
        & & &
        CF-informed & Informed &  \textbf{35.4} &  \textbf{75.0}  & 0.344
        \\
        &&& 
         \highlighter Informed &  \highlighter CF-informed &  \highlighter 25.7 &  \highlighter 65.2 &  \highlighter 0.338
        \\
        &&&
        CF-informed & CF-informed &
        30.0 & 69.2 & 0.414
        \\ \midrule
        \multirow{10}{*}{GPT-4}& \multirow{5}{*}{Human} & \multirow{5}{*}{Human} & 
        
        Vanilla & Vanilla & 35.9 & 52.2 & 0.382   
        \\
        & & & 
         \highlighter Informed &  \highlighter Informed  &  \highlighter 38.1 &  \highlighter 57.0
         &  \highlighter 0.392
        \\
        & & &
        CF-informed & Informed & 35.2  & 52.0 & 0.370
        \\
        &&& 
         \highlighter Informed &  \highlighter CF-informed &  \highlighter \textbf{42.5} &  \highlighter \textbf{61.4}   &  \highlighter \textbf{0.394}
        \\
        &&&
        CF-informed & CF-informed &   36.8 & 55.8
         & 0.382
        
        \\ \cmidrule(r){2-8}
        & \multirow{5}{*}{LLM} & \multirow{5}{*}{LLM} & 
        
        Vanilla & Vanilla & 37.8 & 54.2
         & 0.304
        \\
        & & & 
         \highlighter Informed &  \highlighter Informed &  \highlighter 36.3 &  \highlighter 53.0 &  \highlighter 0.296
        \\
        & & &
        CF-informed & Informed & \textbf{40.5} &  \textbf{58.4}  & 0.298
        \\
        &&& 
         \highlighter Informed &  \highlighter CF-informed &  \highlighter 35.2 &  \highlighter 53.2  &  \highlighter 0.294 
        
        \\
        &&&
        CF-informed & CF-informed & 37.1 & 55.4 
         & 0.294
        
         \\ \bottomrule
    \end{tabular}
    
    \caption{Quality of attribution and answer correctness with Mixed RAG modes for the MS MARCO benchmark. The columns ``Relevant Documents'' and ``Non-relevant Documents'' refer to the actual authorship of input documents.
    }
    \label{tab:mixed_full2}
\end{table*}
\section{Examples}
Table \ref{tab:example_dragon} shows the results of Authorship-Informed retrieval-augmented generation with \texttt{Mistral}, \texttt{Llama3}, and \texttt{GPT-4} for the query ``where was the new pete's dragon filmed.'' Both \texttt{Llama3} and \texttt{GPT-4} generate the correct answer and accurately attribute their answers to the ground-truth document \texttt{[5]}. However, despite providing the correct answer and the correct attribution, \texttt{Mistral} attributes its generated answer to an additional source document, i.e., document \texttt{[2]}. Table \ref{tab:example_australia} shows the results of three RAG modes with \texttt{GPT-4} for the query ``who won the men's single title of australia open on 1 february 2015.'' This result corresponds to the combination of ``human-written'' relevant documents and LLM-written non-relevant ones. As we see, in all RAG models, this model makes a mistake in attributing to document [2], which does not contain the answer. However, in the Authorship Informed RAG mode (where we inform the LLM that document \texttt{[8]} has human authorship), in addition to document \texttt{[2]}, the model also refers to the ground-truth document \texttt{[8]}. 
\begin{table*}[ht]
    \centering
    \fontsize{9}{10}\selectfont
    \ttfamily
    \begin{tabular}{p{15cm}}
        \toprule
        \highlighter
        \textbf{Query} 
        \\
        where was the new pete's dragon filmed
        \\
         \highlighter 
         \textbf{Ground Truth Answer:}
         \\
        New Zealand
        \\ 
        \highlighter \textbf{Groundtruth Document: [5]}
        \\
Pete's Dragon (2016 film). Principal photography commenced in January 2015 in New Zealand, with Lowery directing. Rehearsals began early January 2015, while news reported the beginning of the shooting on January 26, 2015, set to last through April. Live-action filming locations included Bay of Plenty, Taupo and Wellington, while CGI was done at Stone Street Studios. On February 10, 2015, a press release confirmed that principal photography had commenced. Shooting took place in and around Wellington and Rotorua, before transferring to Tapanui, where the old Blue Mountain Lumber mill was used as the mill in the film, and the main street became Millhaven for two weeks. Production concluded on April 30, 2015.
         \\ \midrule
\highlighter \textbf{Mistral Output}
         \\
    "The new Pete's Dragon (2016) was filmed in various locations in \highlightgreen{New Zealand}, including the Bay of Plenty, Taupo, and Wellington \highlightred{[2]}\highlightgreen{[5]}. The production also used the old Blue Mountain Lumber mill in Tapanui as the film's mill and transformed the main street into Millhaven for two weeks \highlightgreen{[5]}."
         \\ \midrule
\highlighter \textbf{Llama3 Output}
         \\
According to Document \highlightgreen{[5]}, the live-action scenes of the 2016 film "Pete\'s Dragon" were shot in various locations in \highlightgreen{New Zealand}, including the Bay of Plenty, Taupo, Wellington, and Tapanui.
        \\ \midrule        
         \highlighter \textbf{GPT-4 Output} 
         \\
The new Pete's Dragon (2016) was filmed in various locations in \highlightgreen{New Zealand}, including the Bay of Plenty, Taupo, Wellington, Rotorua, and Tapanui \highlightgreen{[5]}.
\\ \midrule
\highlighter \textbf{Document [2]} (wrongly cited in Mistral output)
\\ Pete's Dragon (2016 film). In March 2013, Walt Disney Pictures announced it would remake Pete's Dragon, as non-musical, dramatic film to be written by David Lowery and Toby Halbrooks, and directed by Lowery. Disney intended to rework the core story as a dramatic story, rather than a musical. It set the new work in the Pacific Northwest of the early 1980s, rather than the Northern New England of the early 1900s in the 1977 film.
\\

    \\ \bottomrule
    \end{tabular}
    \caption{Answer/Attribution generated by \texttt{Mistral}, \texttt{Llama3}, and \texttt{GPT-4} in Authorship Informed RAG mode with synthetic relevant and human-written (original) non-relevant documents. 
    }
    \label{tab:example_dragon}
\end{table*}
\begin{table*}[ht]
    \centering
    \fontsize{9}{10}\selectfont
    \ttfamily
    \begin{tabular}{p{15cm}}
        \toprule
        \highlighter
        \textbf{Query} 
        \\
        who won the men's single title of australia open on 1 february 2015
        \\
         \highlighter 
         \textbf{Ground Truth Answer:}
         \\
        'Novak Djokovic', 'Djokovic'
        \\ 
        \highlighter \textbf{Groundtruth Document:} [8]
        \\
2015 Australian Open – Men's Singles. Djokovic went on to win his fifth Australian Open title by defeating Murray once again, 7–6(7–5), 6–7(4–7), 6–3, 6–0. The defeat saw Murray lose the Australian Open final for the fourth time in his career, becoming the first man in the Open Era to achieve four runner-up finishes in Melbourne.
        \\ \midrule
         \highlighter \textbf{Vanilla}
         \\
\highlightred{Stan Wawrinka} won the men's singles title at the 2015 Australian Open \highlightred{[2]}.
         \\
         \highlighter \textbf{Informed}
         \\
\highlightred{Stan Wawrinka} won the men's singles title at the 2014 Australian Open, defeating the world's top two players \highlightred{[2]}. However, the 2015 Australian Open men's singles title was won by \highlightgreen{Novak Djokovic}, who defeated Andy Murray in the final \highlightgreen{[8]}.
         \\ \midrule
\highlighter \textbf{CF-Informed}
         \\
\highlightred{Stan Wawrinka} won the men's singles title at the 2015 Australian Open \highlightred{[2]}.
\\ \midrule
\highlighter \textbf{Document [2]} (wrongly cited in Vanilla and CF-Informed RAG Modes)
\\ 
2015 French Open – Men's Singles. Wawrinka defeated Djokovic in the final, 4–6, 6–4, 6–3, 6–4, to win his first French Open and second Grand Slam title. As he did when he won the 2014 Australian Open, Wawrinka defeated the world No. 1 and world No. 2 on his way to the title.
    \\ \bottomrule
    \end{tabular}
    \caption{Answer/Attribution generated by \texttt{GPT-4} in Vanilla, Authorship Informed, and Counterfactual-Authorship Informed RAG modes, with human-written (original) relevant and synthetic non-relevant documents. Reminding LLMs about the authors (Authorship Informed RAG mode) has resulted in a correct answer and attribution.}

    \label{tab:example_australia}
\end{table*}
\end{document}